\documentclass[runningheads]{llncs}

\usepackage{hyperref}

\usepackage{graphicx}
\usepackage{comment}
\usepackage{amsmath,amssymb} %
\usepackage{color}

\usepackage{tabularx}
\usepackage{booktabs} %
\usepackage{subfig}
\usepackage{wrapfig}

\newcommand{\bK}{\mathbf{K}}

\newcommand{\bR}{\mathbf{R}}

\newcommand{\bt}{\mathbf{t}}
\newcommand{\bu}{\mathbf{u}}

\newcommand{\cL}{\mathcal{L}}
\newcommand{\cM}{\mathcal{M}}

\makeatletter
\DeclareRobustCommand\onedot{\futurelet\@let@token\@onedot}
\def\@onedot{\ifx\@let@token.\else.\null\fi\xspace}

\makeatother

\newcommand\myparagraph[1]{\vspace{1mm}\noindent\textbf{#1}}

\definecolor{darkgreen}{rgb}{0,0.7,0}

\begin{document}
\pagestyle{headings}
\mainmatter
\def\ECCVSubNumber{3314}  %

\title{Free View Synthesis} %

\titlerunning{Free View Synthesis}
\author{Gernot Riegler \and Vladlen Koltun}
\authorrunning{G. Riegler and V. Koltun}
\institute{Intel Labs}

\maketitle

\begin{abstract}
We present a method for novel view synthesis from input images that are freely distributed around a scene. Our method does not rely on a regular arrangement of input views, can synthesize images for free camera movement through the scene, and works for general scenes with unconstrained geometric layouts.
We calibrate the input images via SfM and erect a coarse geometric scaffold via MVS. This scaffold is used to create a proxy depth map for a novel view of the scene.
Based on this depth map, a recurrent encoder-decoder network processes reprojected features from nearby views and synthesizes the new view.
Our network does not need to be optimized for a given scene. After training on a dataset, it works in previously unseen environments with no fine-tuning or per-scene optimization.
We evaluate the presented approach on challenging real-world datasets, including Tanks and Temples, where we demonstrate successful view synthesis for the first time and substantially outperform prior and concurrent work.
\keywords{view synthesis, image-based rendering}
\end{abstract}

\section{Introduction}

\begin{figure}[t]
    \centering
    \begin{tabular}{c c}
        \includegraphics[width=0.49\linewidth]{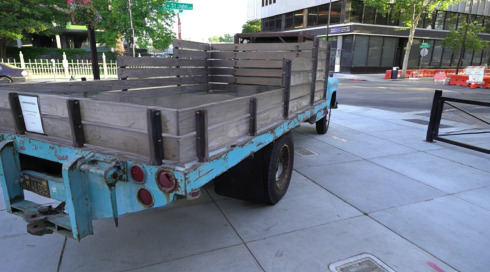} &
        \includegraphics[width=0.49\linewidth]{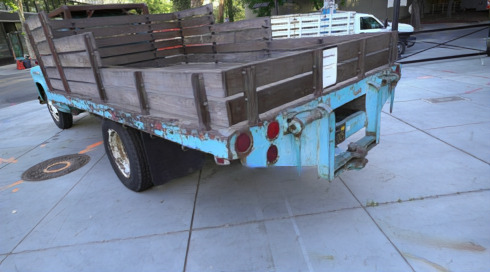} \\
        \includegraphics[width=0.49\linewidth]{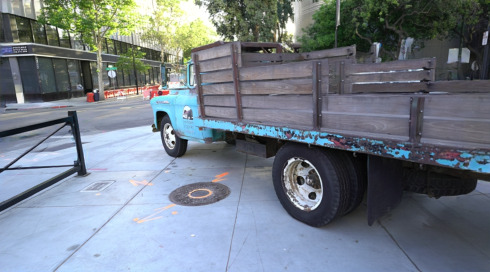} &
        \includegraphics[width=0.49\linewidth]{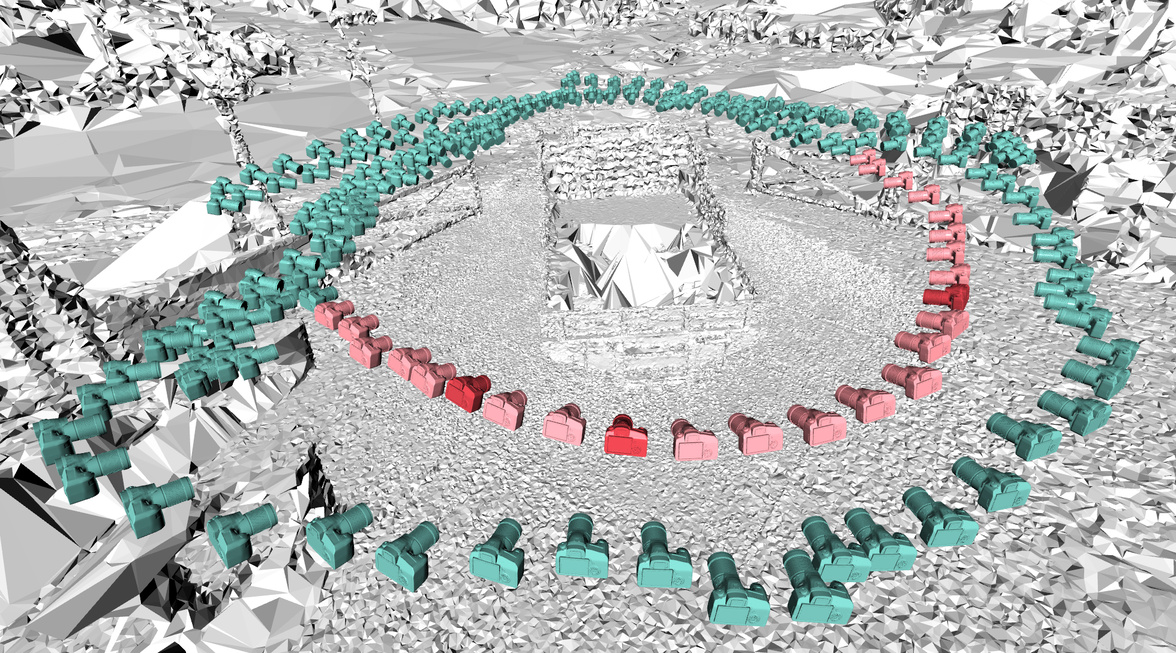}
    \end{tabular}
    \vspace{-1em}
    \caption{
        Novel view synthesis from unstructured input images.
        The first three images show our synthesized results on the \emph{Truck} scene from Tanks and Temples~\cite{Knapitsch2017Tanks}.
        The unstructured image sequence was recorded using a handheld camera in natural motion.
        We repurpose the Tanks and Temples dataset to evaluate view synthesis by using a subset of the images as input (green cameras in the bottom right image).
        The other views, which significantly deviate from the input, act as target poses for view synthesis (red cameras).
    }
    \vspace{-2em}
    \label{fig:teaser}
\end{figure}

Suppose you want to visit the Sagrada Fam{\'i}lia in Barcelona but cannot travel there in person due to a coronavirus pandemic that shut down travel across the globe.
Virtual reality could offer a surrogate for physically being there.
For the experience to be maximally compelling, two requirements must be met.
First, you should be free to move through the scene: you should be able to go anywhere in the environment, freely moving your head and body.
Second, the synthesized images should be photorealistic: perceptually indistinguishable from reality.

In this paper, we present a method for free view synthesis from unstructured input images in general scenes.
Given a set of images or a video of a scene, our approach enables the rendering of a completely new camera path.
See Figure~\ref{fig:teaser} and the supplementary video for examples.
We use 3D proxy geometry to map information from the source images to the novel target view.
Rather than mapping the color values of the source images, we first encode them using a deep convolutional network.
Utilizing the proxy geometry, we map the encoded features from the source images into the target view and blend them via a second network.
Since the target views can deviate significantly from the source views, we develop a recurrent blending network that is invariant to the number of input images.

Experimental results indicate that our approach works much better than state-of-the-art methods across challenging real-world datasets.
On the Tanks and Temples dataset~\cite{Knapitsch2017Tanks}, we reduce the LPIPS error~\cite{Zhang2018Unreasonable} by more than a factor of 2 on all scenes with respect to state-of-the-art methods such as EVS~\cite{Choi2019Extreme} and LLFF~\cite{Mildenhall2019Local}.
On the DTU dataset~\cite{Aanaes2016DTU}, we also significantly reduce LPIPS relative to EVS and LLFF.

Furthermore, we convincingly outperform methods that are published \emph{concurrently} with our work: Neural Radiance Fields (NeRF)~\cite{Mildenhall2020Nerf} and Neural Point-Based Graphics (NPBG)~\cite{Aliev2020Neural}.
We reduce LPIPS relative to these concurrent methods on Tanks and Temples and perform on par on DTU.
We also observe that NPBG performs well on Tanks and Temples and poorly on DTU, NeRF performs well on DTU and poorly on Tanks and Temples, while our approach performs well across datasets.

\section{Related Work}

\myparagraph{Image-based rendering without deep learning.}
Image-based rendering aims to enable the synthesis of new views of a scene directly from a set of input images~\cite{Chen1993View,Debevec1996Modeling,Gortler1996Lumigraph,Heigl1999Plenoptic,Levoy1996Light,Seitz1996ViewMorphing,Shum2000review,Zitnick2004High}.
Different methods map information from the input images to the target view in different ways.
Early light-field methods~\cite{Gortler1996Lumigraph,Levoy1996Light,Seitz1996ViewMorphing} do not require any information about the scene geometry, but either require a fairly dense and regularly-spaced camera grid, or restrict the target view to be a linear combination of the source views.
Heigl et al.~\cite{Heigl1999Plenoptic} compute depth maps per view via stereo matching and use them for view synthesis.
Bilinear blending of viewpoints is possible if the cameras are located approximately on the surface of a sphere and the object is centered~\cite{Davis2012Unstructured}.
These approaches impose restrictions on the layout of the input views, while we target unstructured settings in which the input views are distributed freely around the scene, for example with a single handheld video sequence.

Approaches for unstructured view synthesis are commonly based on constructing 3D proxy geometry that guides the synthesis.
Buehler et al.~\cite{Buehler2001Unstructured} describe the Unstructured Lumigraph, which utilizes dense and accurate 3D geometry to map and blend the input images in a novel target view.
Chaurasia et al.~\cite{Chaurasia2013depthSynthesis} estimate a per-view depth map and use these to map color values and blending weights into the target view.
The method leverages superpixels to make up for missing depth values.
Hyperlapse~\cite{Kopf2014First} also uses 3D proxy geometry obtained via structure-from-motion (SfM) and multi-view stereo (MVS).
To composite a clean image in the target view, the method optimizes a Markov random field.

Rather than estimate depth from input color images, some systems assume that the input views were acquired by an RGB-D sensor that provided dense depth maps of the scene.
Hedman et al.~\cite{Hedman2016Scalable} utilize such an RGB-D sensor to aid their fast rendering pipeline.
Penner and Zhang~\cite{Penner2017Soft3D} use a volumetric approach that associates each voxel with a confidence value which indicates whether the voxel encloses free space or a physical surface.

\myparagraph{Image-based rendering with deep learning.}
Deep learning has come to play an important role in image-based rendering.
Deep networks have been used to blend input images in the target view~\cite{Hedman2018Deep,Thies2020Ignor}, to construct neural scene representations~\cite{Aliev2020Neural,Niemeyer2019Differentiable,Sitzmann2019Deepvoxels,Sitzmann2019Scene,Thies2019Deferred}, and to unify geometry estimation and blending in a single model~\cite{Flynn2019Deepview,Kalantari2016Learning}.

Flynn et al.~\cite{Flynn2016DeepStereo} used a plane-sweep volume~\cite{Collins1996Space} within a network architecture for image-based rendering.
A color branch predicts the color values for each depth plane in the target view and a second branch predicts the probability for a given depth plane.
Kalantari et al.~\cite{Kalantari2016Learning} propose a similar system for a light-field setting: four cameras placed on a plane with the same viewing direction.
Their method also constructs a plane-sweep volume with the four given images and computes mean and standard deviation per plane as features.
Based on these features, one network estimates a disparity map and another reprojects the images and processes them.
Hedman et al.~\cite{Hedman2018Deep} use a deep convolutional network to blend source images that have been warped into the target view.
Given a dense and accurate proxy geometry obtained by two independent MVS methods, four image mosaics are generated and are then fed to a network to estimate blending weights.
Our work is related as we also emphasize the role of the mapping and blending network, but we do not require the construction of input mosaics based on hand-crafted heuristics.
Instead, our method can handle an arbitrary number of input images and we output full color images together with blending weights, which enables a certain degree of inpainting.
Thies et al.~\cite{Thies2020Ignor} extend the ideas of Hedman et al.~\cite{Hedman2018Deep} to better handle view-dependent effects.
They train an additional network per scene that estimates view-dependent effects given a depth map of the target view.
Xu et al.~\cite{Xu2019DeepViewSynthesis} also focus on view-dependent effects, but use a structured setup and directional lighting.

Zhou et al.~\cite{Zhou2018StereoMagnification} introduce a multi-plane image representation that is estimated by a convolutional network for stereo magnification. %
The image is represented over multiple RGB-$\alpha$ planes, where each plane is related to a certain depth.
Given this representation, new views can be rendered using back-to-front composition.
Choi et al.~\cite{Choi2019Extreme} build upon MVSNet~\cite{Yao2018Mvsnet} for view extrapolation.
The method estimates a depth probability volume for each input view that is then warped and fused into the target view.
From the fused volume an initial novel view is synthesized.
This is then further refined by comparing and integrating candidate patches from the source images.
Similarly, the method of Srinivasan et al.~\cite{Srinivasan2019PushingBoundaries} synthesizes views from a narrow-baseline pair of images.
The work extends the idea of multi-plane images~\cite{Zhou2018StereoMagnification} and shows the relation between the range of views that can be rendered from a multi-plane image and the depth plane sampling frequency.
Mildenhall et al.~\cite{Mildenhall2019Local} further improve this method with practical user guidance and refined network architectures together with local layered scene representations.
Flynn et al.~\cite{Flynn2019Deepview} considerably improve the view synthesis quality of light-field setups.
They use plane sweep volume inputs~\cite{Collins1996Space} and multi-plane image outputs~\cite{Zhou2018StereoMagnification} together with a network based on regularized gradient descent to gradually refine the generated images.

Instead of mapping features from source images to novel target views, some very recent methods train neural scene representations.
In a work that is published concurrently with ours, Aliev et al.~\cite{Aliev2020Neural} describe Neural Point-Based Graphics, where each 3D point is associated with a learned feature vector.
These features are splatted into the target view and translated via a rendering network to synthesize the output image.
The feature vectors are optimized per scene: application to a new scene requires training the feature extractor for that scene.
Thies et al.~\cite{Thies2019Deferred} use a mesh instead of 3D points to embed the feature vectors.
Sitzmann et al.~\cite{Sitzmann2019Deepvoxels} avoid explicit proxy geometry and project source images into a neural voxel grid, where each voxel is associated with a trainable feature vector.
This representation is likewise trained for each object it is applied to.
Lombardi et al.~\cite{Lombardi2019NeuralVolumes} avoid memory-intensive high-resolution neural voxel grids by computing warp fields.
In another concurrent work, Mildenhall et al.~\cite{Mildenhall2020Nerf} represent the 5D radiance field by an MLP that can be queried in a volume rendering framework to synthesize new views.
In all these methods, dedicated per-scene training is required to apply the representation to a new scene.
We train our image encoding and blending networks only once on a training set and apply them to new scenes without any per-scene adaptation or fine-tuning.

In a related line of work, new views are synthesized from a single input image~\cite{Niklaus2019KenBurns,Wiles2020SynSin}.
These approaches only allow small deviations from the initial viewpoint, rather than the unrestricted travel through the scene that motivates our work.

\section{Method}
Our method begins with a preprocessing stage that involves estimating the poses of the input images and computing 3D proxy geometry via multi-view stereo and meshing.
Given a target view, we select nearby source images, map them into the target view, and blend them using a recurrent convolutional network.
We now describe each step in detail.

\subsection{Preprocessing}

\myparagraph{Pose estimation.} \label{sec:pose_registration}
Our input is a set of $N$ images $\{I_n\}_{n=1}^N$, for example from a handheld video of a scene.
We begin by estimating the poses of these images.
For this purpose, we rely on well-established structure-from-motion (SfM) techniques.
Specifically, we utilize COLMAP~\cite{Schoenberger2016SfM} to compute camera poses and a sparse 3D point cloud of the scene.
The poses are represented by rotation matrices $\{\bR_n\}_{n=1}^N$ and translation vectors $\{\bt_n\}_{n=1}^N$.
The SfM pipeline also estimates the intrinsic parameters of the cameras $\{\bK_n\}_{n=1}^N$ and distortion coefficients.\footnote{In the case of video input data we assume that the intrinsics are shared for all views. $\forall i \neq j: \bK_i = \bK_j$.}
We use these to undistort all images.
In the remainder of the paper, the set of images~$\{I_n\}_{n=1}^N$ refers to the set of undistorted images.

\myparagraph{Proxy geometry.} \label{sec:proxy_geometry}
For the mapping of the source images to the target view and also for the selection of the source images that are used to synthesize a novel view $\hat{I}_t$, we need 3D proxy geometry $\cM$.
We run a standard multi-view stereo (MVS) method~\cite{Schoenberger2016Pixelwise} to estimate a depth map for each source image.
The depth maps are further fused into a coherent 3D point cloud using the fusion algorithm implemented in COLMAP~\cite{Schoenberger2016Pixelwise,Schoenberger2016SfM}.
We also experimented with more recent MVS methods~\cite{Yao2018Mvsnet,Yao2019Recurrent}, but did not observe significant improvements in the mapping and blending performance on large scale scenes.

\begin{figure}[t]
    \centering
    \subfloat[][Point cloud]{
        \includegraphics[width=0.49\textwidth]{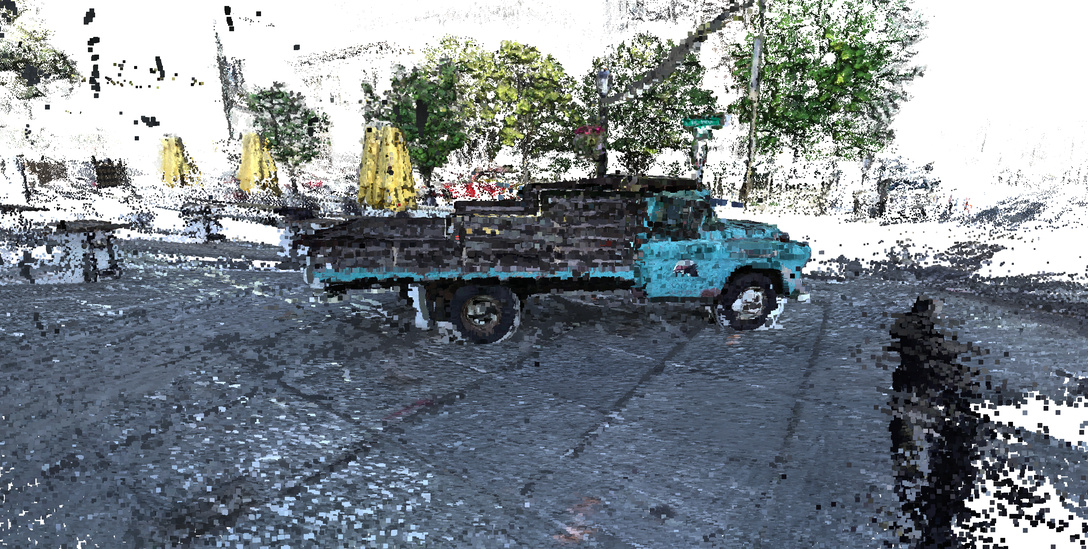}
    }
    \subfloat[][Mesh]{
        \includegraphics[width=0.49\textwidth]{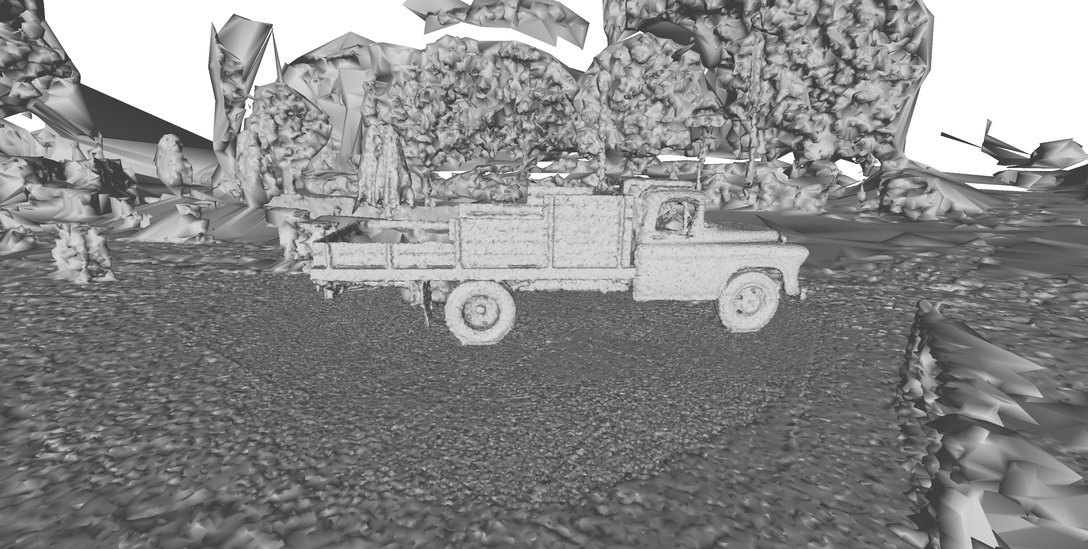}
    }
    \vspace{-2mm}
    \caption{
        Proxy geometry for view synthesis.
        We use a surface mesh extracted from a Delaunay tetrahedralization (right).
        While it is more complete than the point cloud from MVS (left), it also introduces spurious triangles.
    }
    \vspace{-4mm}
    \label{fig:proxy_geometry}
\end{figure}

For the mapping of the source image features, we rely on a depth map in the novel target view $D_t$ that is derived from the proxy geometry.
However, rendering depth maps from 3D point clouds is problematic for two reasons.
First, 3D points that are in the background and should be occluded by a surface can be projected into the foreground, leading to invalid depth values. Second, in larger untextured regions it is likely that no 3D points are reconstructed.
Both problems can be alleviated by fitting a surface mesh to the 3D point cloud.
We utilize a Delaunay-based reconstruction~\cite{Jancosek2011Multi,Labatut2007Efficient} as it can tolerate a certain amount of outliers.
The roughness of the resulting surface can be handled by our subsequent blending network.
The resulting surface mesh $\cM$ is used to derive depth maps of the source views $\{D_n\}_{n=1}^N$ and of any target view $D_t$.
Figure~\ref{fig:proxy_geometry} shows our computed proxy geometry for a Tanks and Temples scene~\cite{Knapitsch2017Tanks}.

\subsection{Selection of Source Images}

A core advantage of the mapping and blending network described in the next section is that it supports an arbitrary number of source input images.
However, in practical situations, many source images will have no overlap with the target view, because they are facing into the opposite direction, or are taken from a very different location.
Also during network training, we are limited by GPU memory and can only use a fixed number of input images.
For these reasons, we have a simple but effective source selection strategy.
Based on the proxy geometry $\cM$ we select the $K$ out of $N$ source images that maximize the overlap with the target view.

Specifically, we derive for each target view a depth map $D_t$.
Each pixel of the depth map $D_t$ with a valid depth value is projected into the domains of all source images based on the user defined intrinsic and extrinsic parameters of the target view $\bK_t$, $\bR_t$, and $\bt_t$ and the estimated intrinsic and extrinsic parameters of the source images $\{\bK_n\}_{n=1}^N$, $\{\bR_n\}_{n=1}^N$, and $\{\bt_n\}_{n=1}^N$.
For each source image, we count the number of pixels from the target view that are mapped to the valid source image domain and select the top $K$ images that maximize that score.
To further handle occlusions and other outliers in this process, we only count pixels where the projected target depth is within $1\%$ of the source depth.

\subsection{Mapping and Blending}

\begin{figure}[t]
    \centering
    \includegraphics[width=0.9\linewidth]{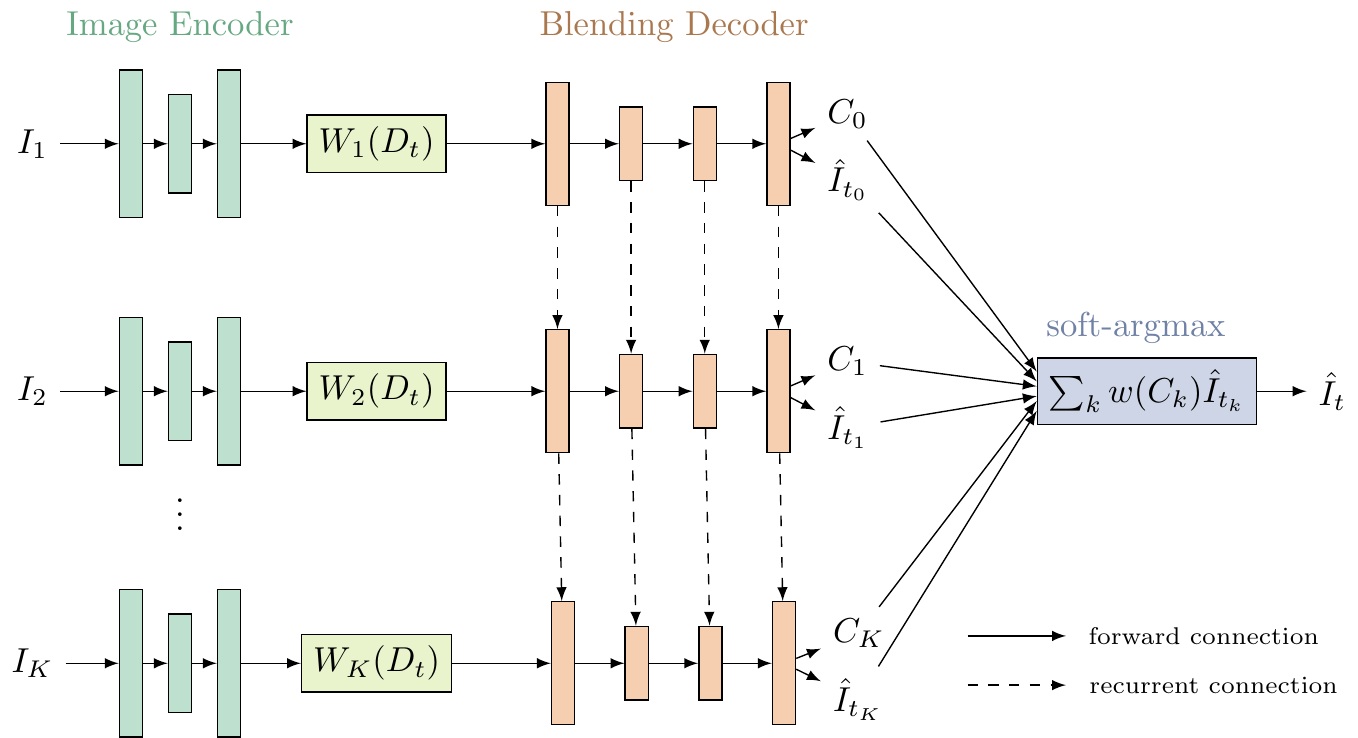}
    \vspace{1mm}
    \caption{
        Overview of the recurrent mapping and blending network.
        The input is a set of source images $\{I_k\}_{k=1}^K$ that are first encoded by a shared convolutional network, the \emph{image encoder}.
        We map the resulting features into the target view using the depth map $D_t$, derived from the proxy geometry $\cM$.
        The features are then aggregated by a recurrent network, the \emph{blending decoder}.
        For each input image $I_k$, we output a confidence image $C_k$ and a color image $\hat{I}_{t_k}$, which are then aggregated to a final output $\hat{I}_t$ via a soft-argmax using $w(C_k) = \exp(C_k) / \sum_j \exp(C_j)$.
    }
    \vspace{-4mm}
    \label{fig:arch}
\end{figure}

Once we have selected the source images $\{I_k\}_{k=1}^K$, we need to map them into the novel target view and blend the information to an output image $\hat{I}_t$.
For this purpose, we have developed a recurrent mapping and blending network.
We first encode each source image via a U-Net based convolutional network~\cite{Ronneberger2015UNet}.
The features are then mapped into the novel target view and sequentially processed by a blending network that is based on convolutional gated recurrent units (GRU)~\cite{Cho2014GRU}.
For each source image $I_k$, the blending network outputs per-pixel confidence values and a color image in the target view.
The final image $\hat{I}_t$ is then generated by a soft-argmax over these confidence values and color images.
Figure~\ref{fig:arch} provides an overview of the recurrent mapping and blending network.

\myparagraph{Encoding source images.}
In the first part of the mapping and blending network, we encode each source image $I_k$ with a U-Net based convolutional network.
The encoder of the U-Net consists of the first three stages of an ImageNet pretrained VGG network~\cite{Simonyan2015Very} where we replaced the max-pooling with average pooling layers.
In the decoder of the U-Net, we upsample the output features of the previous stage using nearest-neighbor interpolation, concatenate them with the encoder output of the same resolution, and process this by two additional convolutional layers.
All convolutional layers are followed by a ReLU~\cite{Nair2010Rectified}.

\myparagraph{Mapping into the target view.}
The encoded source images must then be mapped into the target view.
For this, we rely on the depth map in the target view $D_t$ that is derived from the proxy geometry $\cM$.
We can gather the feature vectors from the source views via a warping operation $W_k(D_t)$.
For a pixel in homogeneous coordinates $\bu_t$ in the target view, we select the feature vector in the source image $k$ at the location $\bu_k = \bK_k (\bR_r D_t(\bu_t) \bK_t^{-1} \bu_t + \bt_r)$, with relative rotation $\bR_r = \bR_k \bR_t^T$ and relative translation $\bt_r  = \bt_s - \bR_r \bt_t$.
As the locations $\bu_k$ will not be located at the center of the pixel in general, we bilinearly interpolate the features in the warping.
Further, in several cases $\bu_k$ will not be inside the source image domain.
In those instances, we set the warped feature to zero and additionally indicate those locations in a mask that is concatenated to the warped features.
A major problem with the proxy geometry is that several areas do not have associated depth values, especially structures that are far away, or the sky.
To alleviate this problem, we warp features that do not have a valid depth value associated using $+\infty$ as depth value and also concatenate a mask that indicates those features.
This greatly reduces artifacts in the background.

\myparagraph{Blending.}
At this point, we have the feature maps of $K$ source images warped into the target view and now we need to aggregate the information to obtain a single blended output image $I_t$.
A suitable network structure for this kind of problem is a recurrent architecture.
More specifically, we utilize a U-Net based convolutional architecture with gated recurrent units (GRU)~\cite{Cho2014GRU} that regularizes and blends the source feature maps along the spatial dimensions and across the source views.
In principle, all convolutional layers of the blending network can be replaced by a convolutional GRU.
However, we observed that just replacing the last convolutional layer per stage in the encoder, and the first convolutional layer per stage in the decoder works as well and decreases the time needed for training and evaluation of the overall network.
For each source image $I_k$, the blending decoder outputs confidence values $C_k$ and color information $\hat{I}_{t_k}$ per pixel in the target view.
The final output $\hat{I}_t$ is then generated by a soft-argmax $\hat{I}_t = \sum_k w(C_k) \hat{I}_{t_k}$, where $w$ are weights computed by a softmax over the confidence values.

\myparagraph{Training.}
To train the mapping and blending network we require a supervision signal.
We use a natural setup~\cite{Flynn2019Deepview,Hedman2018Deep,Thies2020Ignor}: sample one of the source images, withhold it, and use it as ground-truth $I_t$.

As training loss we utilize the perceptual loss of Chen and Koltun~\cite{Chen2017Photographic}.
Given our estimated image $\hat{I}_t$ and the ground-truth target $I_t$, the loss is
\begin{equation}
    \cL(\hat{I}_t, I_t) = ||\hat{I}_t - I_t||_1 + \sum_l \lambda_l ||\phi_l(\hat{I}_t) - \phi_l(I_t)||_1\,,
\end{equation}
where $\phi_l$ are the outputs of the layers `conv1\_2', `conv2\_2', `conv3\_2', `conv4\_2', and `conv5\_2' of a pretrained VGG-19 network~\cite{Simonyan2015Very}.
The weighting coefficients $\{ \lambda_l \}_{l=1}^5$ are set as in \cite{Chen2017Photographic}.

We use ADAM~\cite{Kingma2014Adam} with a learning rate of $10^{-4}$ and set $\beta_1 = 0.9$, $\beta_2 = 0.9999$, and $\epsilon = 10^{-8}$ to train the recurrent mapping and blending network.
We train the model for 450,000 iterations with a batch size of 1.

\myparagraph{Acceleration.}
The recurrent nature of our mapping and blending network allows the integration of an arbitrary number of source images with a low memory footprint.
To further speed up processing, we precompute the feature embeddings of the source images.
This avoids the repeated encoding of the source images for different synthesized views.

\section{Experimental Evaluation}
We first evaluate our design choices in controlled experiments and then compare to the state of the art.
For each scene, we first run the COLMAP SfM pipeline~\cite{Schoenberger2016SfM} to get camera poses and a sparse reconstruction as described in Section~\ref{sec:pose_registration}.
We also create a dense reconstruction of all models using MVS~\cite{Schoenberger2016Pixelwise} and Delaunay-based surface reconstruction~\cite{Jancosek2011Multi,Labatut2007Efficient} as outlined in Section~\ref{sec:proxy_geometry}.
To train the network we use the Tanks and Temples dataset~\cite{Knapitsch2017Tanks}.
We select $17$ of the $21$ scenes in the dataset for training and supervise the model by designating one image as the ground-truth target and using the remaining ones as source images.
For testing we use scenes that are not included in our training set: \emph{Truck}, \emph{Train}, \emph{M60}, and \emph{Playground}. We chose these scenes for evaluation because the camera paths in these scenes were amenable to extraction of longer subsequences that can be withheld to evaluate significant deviations from the source images.
Note that none of the images from the evaluation scenes have been seen during the training of our method.
See Figure~\ref{fig:teaser} for a visualization of the target and source cameras for the \emph{Truck} scene.
For training, we downsample the images by scaling the image height and width by a factor of four each.
We implemented our recurrent mapping and blending network in PyTorch~\cite{Paszke2019Pytorch}.

In all of our evaluations, we report three different image metrics.
We include PSNR and SSIM to evaluate low-level image differences.
However, those metrics have only weak correlation with human perception~\cite{Zhang2018Unreasonable}.
Therefore, we also include the LPIPS metric, which is based on perceptual features in trained convolutional networks and was shown to better correlate with human perception~\cite{Zhang2018Unreasonable}.

\myparagraph{Architectural choices.}
In the first set of experiments we evaluate our architectural design choices.
See Table~\ref{tab:tat_ablation} for an overview of the results on the Tanks and Temples test scenes.
For these experiments, we also use the quarter resolution images for evaluation.
This evaluation is conducted in the leave-one-out setting, i.e., we select each image per scene once as the unseen ground-truth target and utilize the other images as source images.

\begin{table}[t]
    \centering
    \caption{
        Evaluation of architectural choices on the Tanks and Temples dataset. (Leave-one-out protocol.) See the text for a detailed description of the conditions.
    }
    \resizebox{1\linewidth}{!}{
    \scriptsize
    \begin{tabular}{rrrrrrrrrrrrr}
\toprule
\multicolumn{1}{c}{} & \multicolumn{3}{c}{Truck} & \multicolumn{3}{c}{Train} & \multicolumn{3}{c}{M60} & \multicolumn{3}{c}{Playground} \\
\cmidrule(lr){2-4}
\cmidrule(lr){5-7}
\cmidrule(lr){8-10}
\cmidrule(lr){11-13}
 & $\downarrow$LPIPS & $\uparrow$SSIM & $\uparrow$PSNR & $\downarrow$LPIPS & $\uparrow$SSIM & $\uparrow$PSNR & $\downarrow$LPIPS & $\uparrow$SSIM & $\uparrow$PSNR & $\downarrow$LPIPS & $\uparrow$SSIM & $\uparrow$PSNR \\
\midrule
\multicolumn{1}{l}{{\color[rgb]{0,0,0} Fixed Identity}} & 0.116 & 0.819 & 21.22 & 0.201 & 0.751 & 18.53 & 0.110 & 0.871 & 22.67 & 0.119 & 0.824 & 22.38 \\
\multicolumn{1}{l}{{\color[rgb]{0,0,0} Fixed Encoding}} & 0.096 & 0.828 & 21.19 & 0.168 & 0.769 & 19.01 & 0.096 & 0.876 & 22.80 & 0.107 & 0.831 & 22.40 \\
\multicolumn{1}{l}{{\color[rgb]{0,0,0} Cat Global Avg.}} & 0.089 & 0.842 & 21.49 & 0.175 & 0.773 & 18.73 & 0.093 & 0.887 & 23.41 & 0.098 & 0.845 & 22.92 \\
\multicolumn{1}{l}{{\color[rgb]{0,0,0} Ours w/o Encoding}} & 0.093 & 0.849 & {\bf 22.13} & 0.174 & 0.778 & 19.33 & 0.094 & 0.887 & 23.79 & 0.099 & 0.851 & 23.45 \\
\multicolumn{1}{l}{{\color[rgb]{0,0,0} Ours w/o GRU}} & 0.094 & 0.845 & 21.74 & 0.159 & 0.782 & 19.26 & 0.087 & 0.893 & 23.49 & 0.095 & 0.849 & 23.30 \\
\multicolumn{1}{l}{{\color[rgb]{0,0,0} Ours w/o Masks}} & 0.087 & 0.847 & 21.58 & 0.152 & 0.784 & 19.42 & 0.082 & {\bf 0.897} & 24.07 & 0.087 & 0.850 & 23.16 \\
\multicolumn{1}{l}{{\color[rgb]{0,0,0} Ours w/o inf. depth}} & 0.093 & 0.847 & 21.94 & 0.169 & 0.782 & 18.96 & 0.087 & 0.896 & {\bf 24.08} & 0.094 & 0.853 & 23.47 \\
\multicolumn{1}{l}{{\color[rgb]{0,0,0} Ours w/o soft-argmax}} & 0.091 & 0.845 & 21.74 & 0.159 & 0.786 & 19.43 & 0.086 & 0.891 & 23.79 & 0.090 & 0.857 & 23.50 \\
\multicolumn{1}{l}{{\color[rgb]{0,0,0} Ours full}} & {\bf 0.082} & {\bf 0.852} & 22.03 & {\bf 0.147} & {\bf 0.794} & {\bf 19.54} & {\bf 0.081} & 0.894 & 23.98 & {\bf 0.084} & {\bf 0.859} & {\bf 23.51} \\
\bottomrule
\end{tabular}

    }
    \vspace{-2em}
    \label{tab:tat_ablation}
\end{table}

Our first baseline, \emph{Fixed Identity}, is a network that is related to the one presented in~\cite{Hedman2018Deep}.
It is a U-Net architecture with the same capacity as our blending network, but it receives as input a fixed number ($K=4$) of mapped source images concatenated along the channel dimension and directly outputs the image in the target view.
We use the same source image selection strategy as in our method.
\emph{Fixed Encoding} differs in that we use the same VGG-19 based encoding network prior to mapping as in our approach.
\emph{Cat Global Avg.} is the same architecture as our proposed one without recurrent units, but a global average concatenated to each blending head.
We also ablate our recurrent mapping and blending architecture.
\emph{Ours w/o GRU} uses no GRU in the blending decoder.
In \emph{Ours w/o Encoding} we directly map the input images to the recurrent blending network, and in \emph{Ours w/o Masks} we do not append the mapping masks to the blending network input.
\emph{Ours w/o inf.~depth} does not set the invalid depth values in $D_t$ to $+\infty$, and \emph{Ours w/o soft-argmax} uses a single output head after the evaluation of the last blending decoder instead of the soft-argmax.

The results presented in Table~\ref{tab:tat_ablation} validate the design choices of our method.
A clear advantage is the encoding of the source images before mapping and blending.
We see an overall improvement for the fixed input architecture and our recurrent mapping and blending network.
The difference between our results and the results of \emph{Ours w/o GRU} and \emph{Cat Global Avg.} also highlights the benefit of the recurrent network, i.e., propagating blending information between source images via a recurrent unit.

\begin{wrapfigure}[12]{r}{0.3\textwidth}
    \centering
    \vspace{-3mm}
    \includegraphics[width=\linewidth]{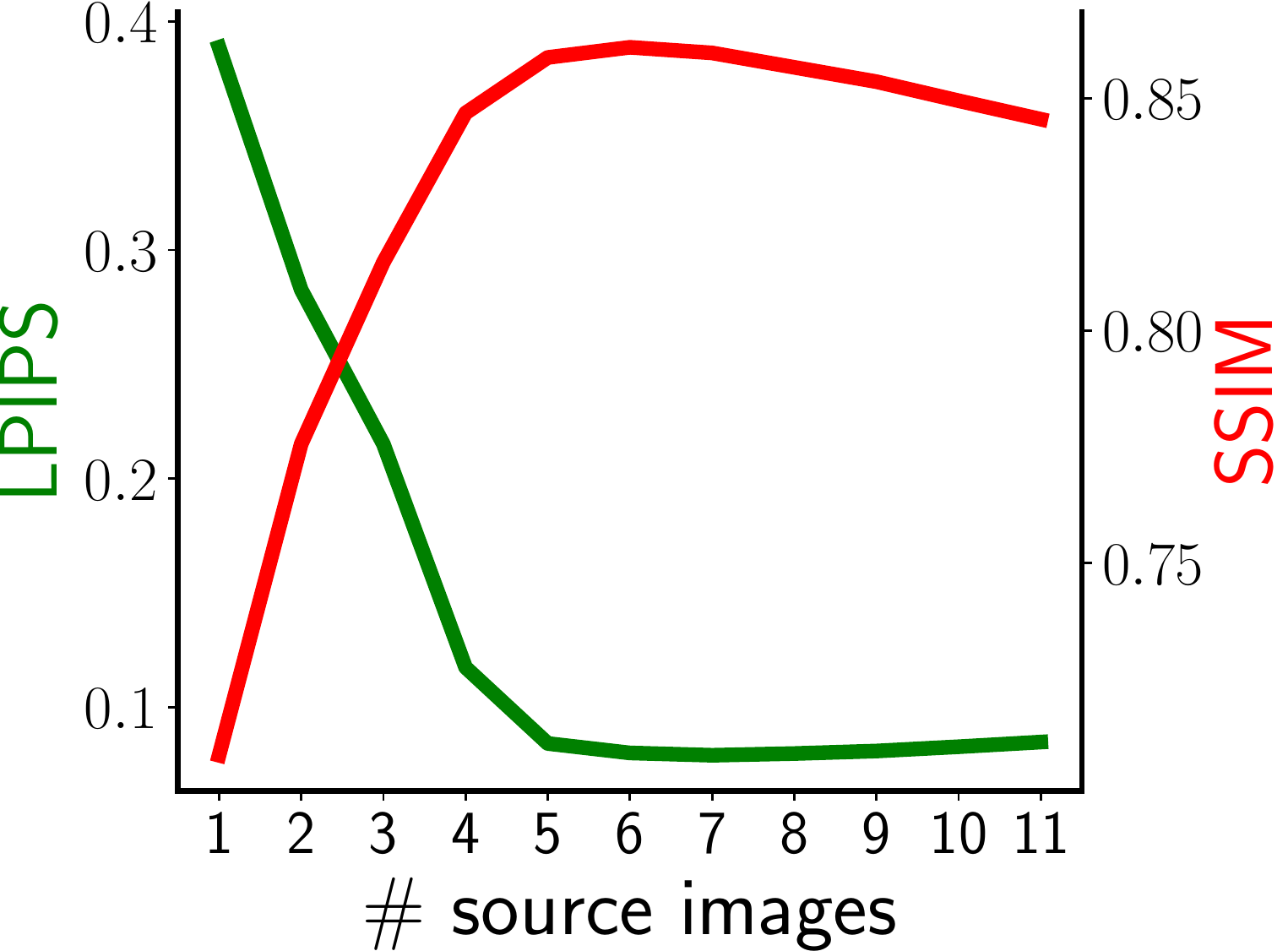}
    \caption{
        The effect of increasing the number of source views.
    }
    \label{fig:tat_n_views}
\end{wrapfigure}
In Figure~\ref{fig:tat_n_views} we evaluate the performance of our method with an increasing number of source images.
We see that image fidelity improves with the number of source images up to 7 images and then saturates.
Note that a higher number of source images is especially important if the novel view is farther away from the scene or object than any of the source images.
In this setting, more source images are needed to cover the view frustum.

\begin{figure}[!t]
    \centering
    \captionof{table}{
        Results on Tanks and Temples. (Whole sequences withheld.)
    }
    \vspace{-2mm}
    \resizebox{1\linewidth}{!}{
    \scriptsize
    \begin{tabular}{rrrrrrrrrrrrr}
\toprule
\multicolumn{1}{c}{} & \multicolumn{3}{c}{Truck} & \multicolumn{3}{c}{Train} & \multicolumn{3}{c}{M60} & \multicolumn{3}{c}{Playground} \\
\cmidrule(lr){2-4}
\cmidrule(lr){5-7}
\cmidrule(lr){8-10}
\cmidrule(lr){11-13}
 & $\downarrow$LPIPS & $\uparrow$SSIM & $\uparrow$PSNR & $\downarrow$LPIPS & $\uparrow$SSIM & $\uparrow$PSNR & $\downarrow$LPIPS & $\uparrow$SSIM & $\uparrow$PSNR & $\downarrow$LPIPS & $\uparrow$SSIM & $\uparrow$PSNR \\
\midrule
\multicolumn{1}{l}{{\color[rgb]{0,0,0} EVS \cite{Choi2019Extreme}}} & 0.41 & 0.563 & 14.99 & 0.64 & 0.454 & 11.81 & 0.62 & 0.473 & 9.66 & 0.39 & 0.610 & 16.34 \\
\multicolumn{1}{l}{{\color[rgb]{0,0,0} LLFF \cite{Mildenhall2019Local}}} & 0.61 & 0.432 & 10.66 & 0.70 & 0.356 & 8.88 & 0.69 & 0.427 & 8.98 & 0.56 & 0.517 & 13.27 \\
\multicolumn{1}{l}{{\color[rgb]{0,0,0} NeRF \cite{Mildenhall2020Nerf}}} & 0.61 & 0.690 & 19.47 & 0.74 & 0.532 & 13.16 & 0.62 & 0.691 & 15.99 & 0.54 & 0.734 & 21.16 \\
\multicolumn{1}{l}{{\color[rgb]{0,0,0} NPBG \cite{Aliev2020Neural}}} & 0.22 & 0.822 & 20.32 & 0.25 & {\bf 0.801} & {\bf 18.08} & 0.36 & 0.716 & 12.35 & 0.17 & {\bf 0.876} & {\bf 23.03} \\
\multicolumn{1}{l}{{\color[rgb]{0,0,0} Our}} & {\bf 0.11} & {\bf 0.867} & {\bf 22.62} & {\bf 0.22} & 0.758 & 17.90 & {\bf 0.29} & {\bf 0.785} & {\bf 17.14} & {\bf 0.16} & 0.837 & 22.03 \\
\bottomrule
\end{tabular}
    }
    \label{tab:tat_seq}
\vspace{6mm}

    \centering
    \begingroup
\setlength{\tabcolsep}{-0.1em}
\renewcommand{\arraystretch}{1}
\begin{tabular}{c c c c c c}
\rotatebox[origin=c]{90}{GT} & \raisebox{-0.5\height}{ \includegraphics[width=0.3167\linewidth]{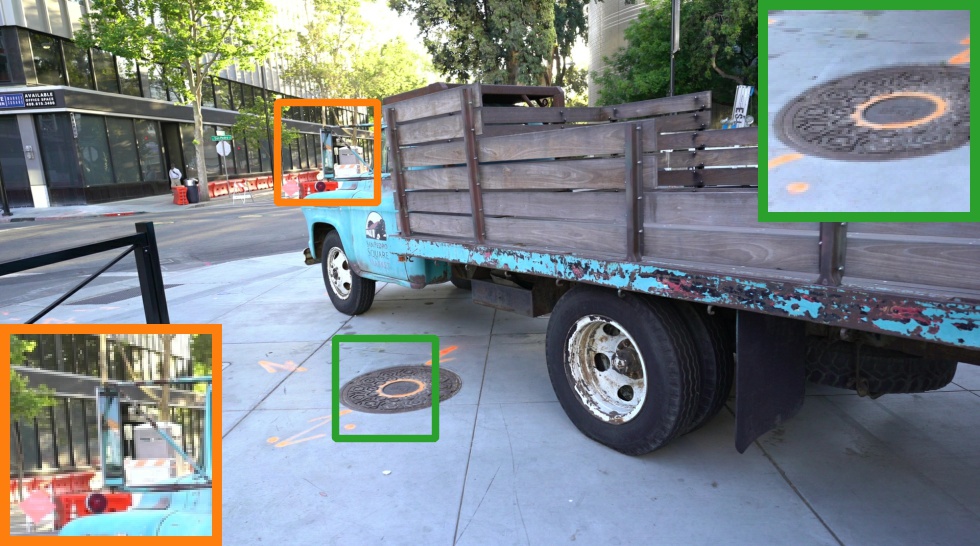} } & \raisebox{-0.5\height}{ \includegraphics[width=0.3167\linewidth]{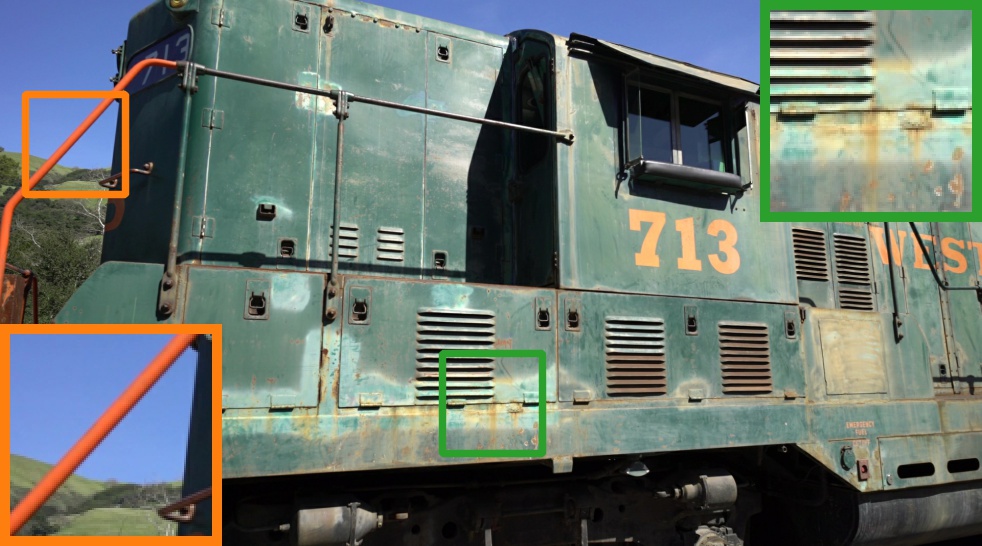} } & \raisebox{-0.5\height}{ \includegraphics[width=0.3167\linewidth]{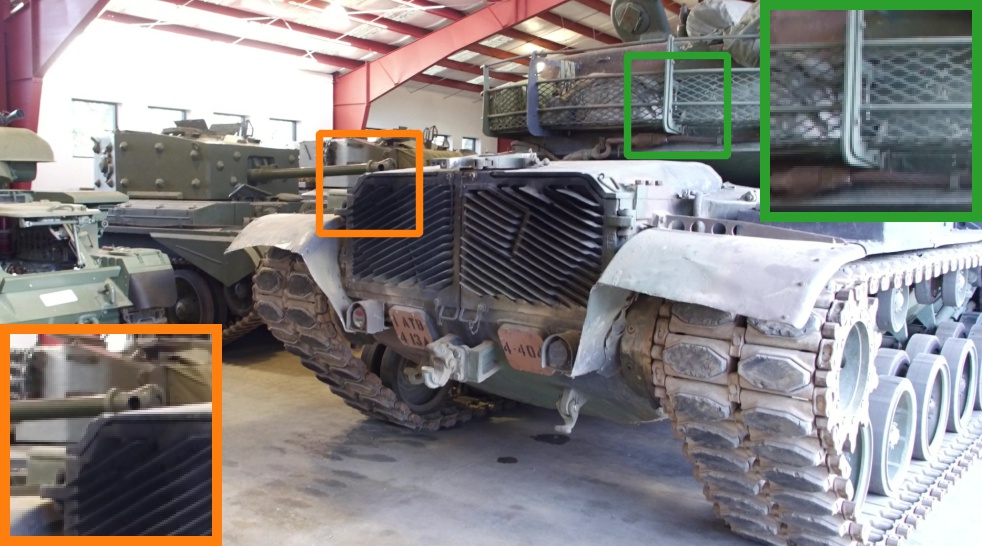} }\vspace{0.25em}\\
\rotatebox[origin=c]{90}{EVS \cite{Choi2019Extreme}} & \raisebox{-0.5\height}{ \includegraphics[width=0.3167\linewidth]{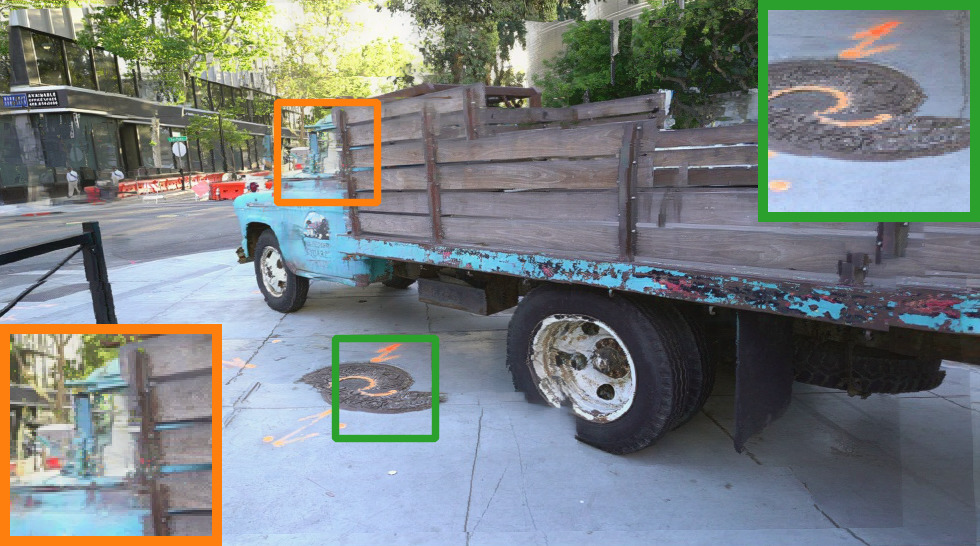} } & \raisebox{-0.5\height}{ \includegraphics[width=0.3167\linewidth]{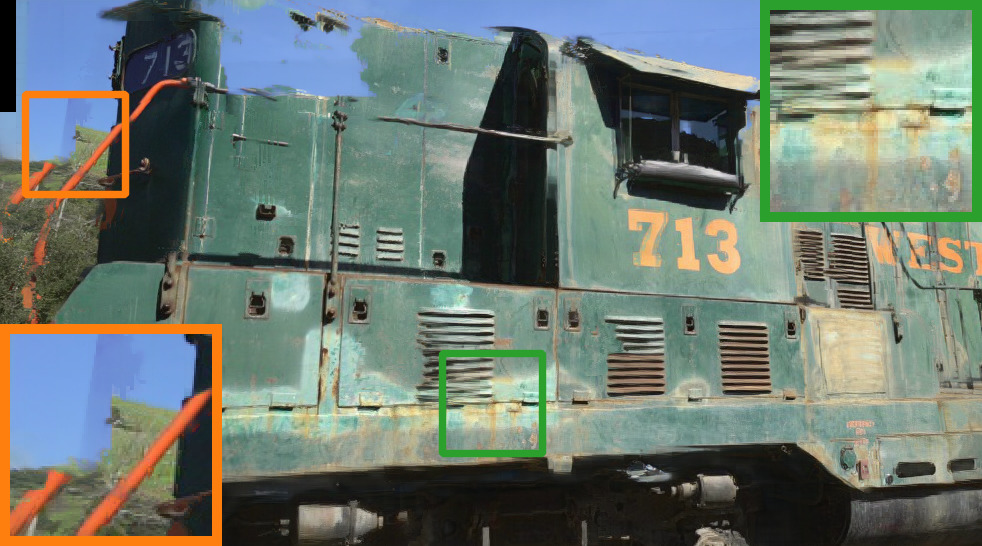} } & \raisebox{-0.5\height}{ \includegraphics[width=0.3167\linewidth]{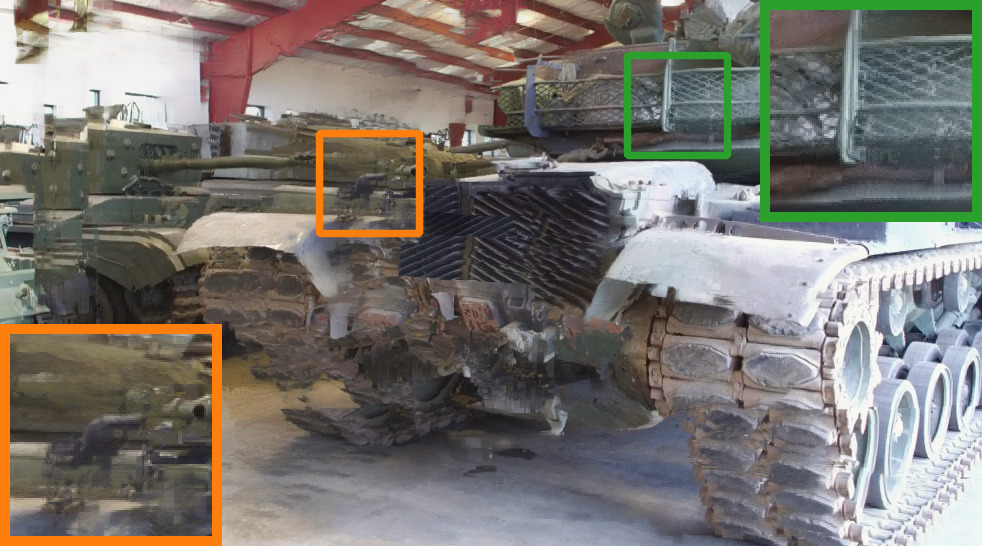} }\vspace{0.25em}\\
\rotatebox[origin=c]{90}{LLFF \cite{Mildenhall2019Local}} & \raisebox{-0.5\height}{ \includegraphics[width=0.3167\linewidth]{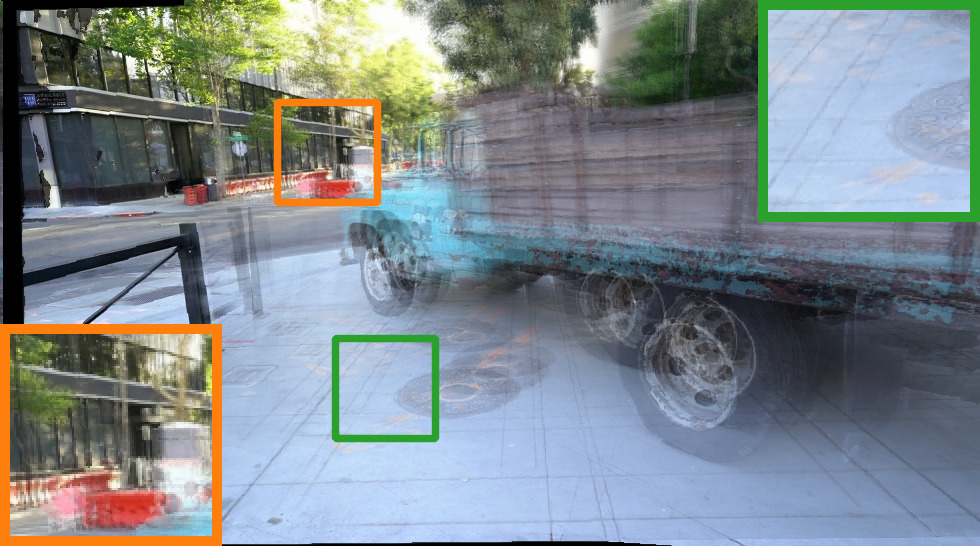} } & \raisebox{-0.5\height}{ \includegraphics[width=0.3167\linewidth]{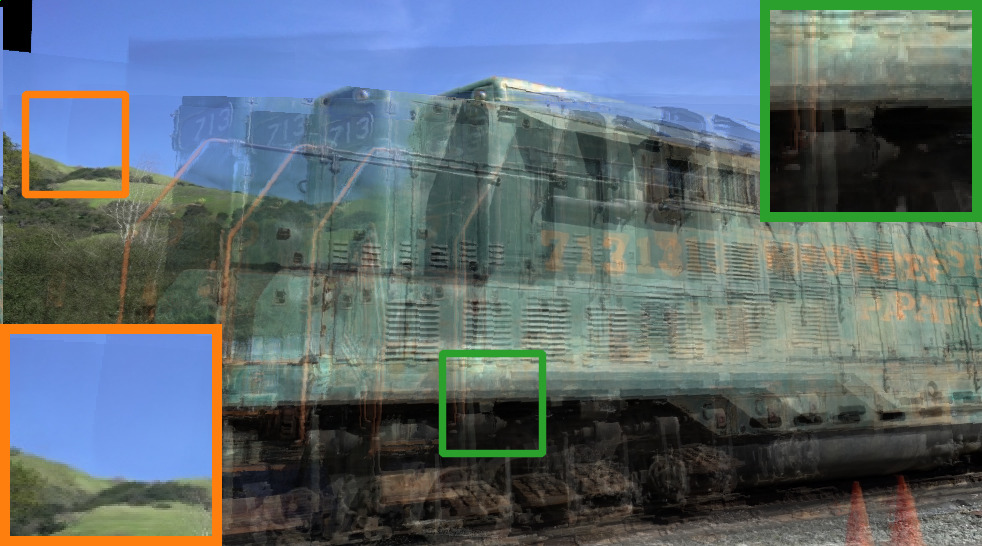} } & \raisebox{-0.5\height}{ \includegraphics[width=0.3167\linewidth]{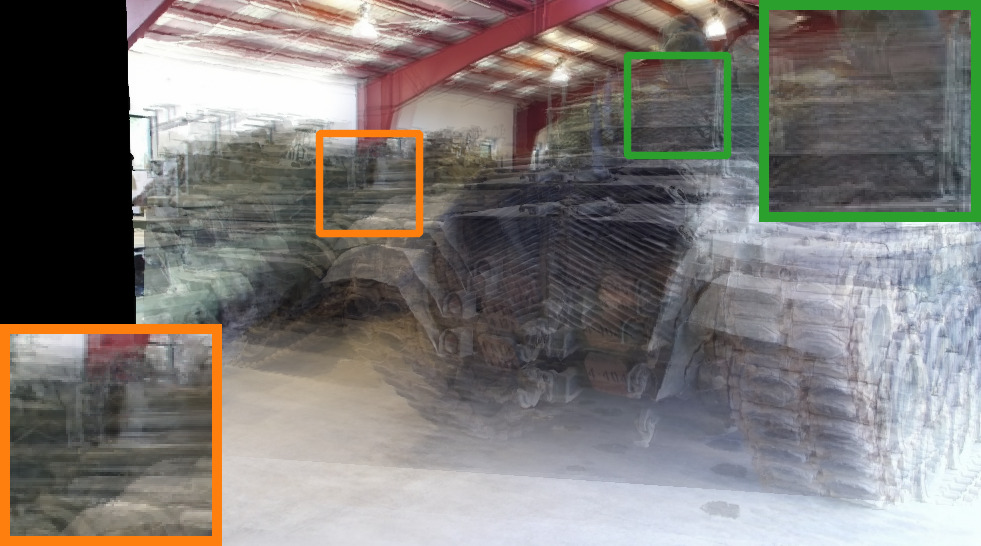} }\vspace{0.25em}\\
\rotatebox[origin=c]{90}{NeRF \cite{Mildenhall2020Nerf}} & \raisebox{-0.5\height}{ \includegraphics[width=0.3167\linewidth]{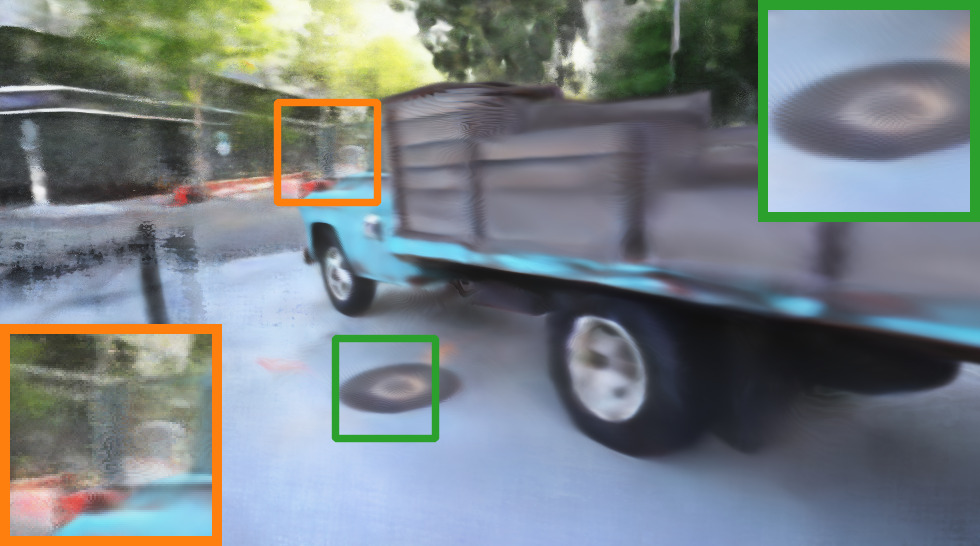} } & \raisebox{-0.5\height}{ \includegraphics[width=0.3167\linewidth]{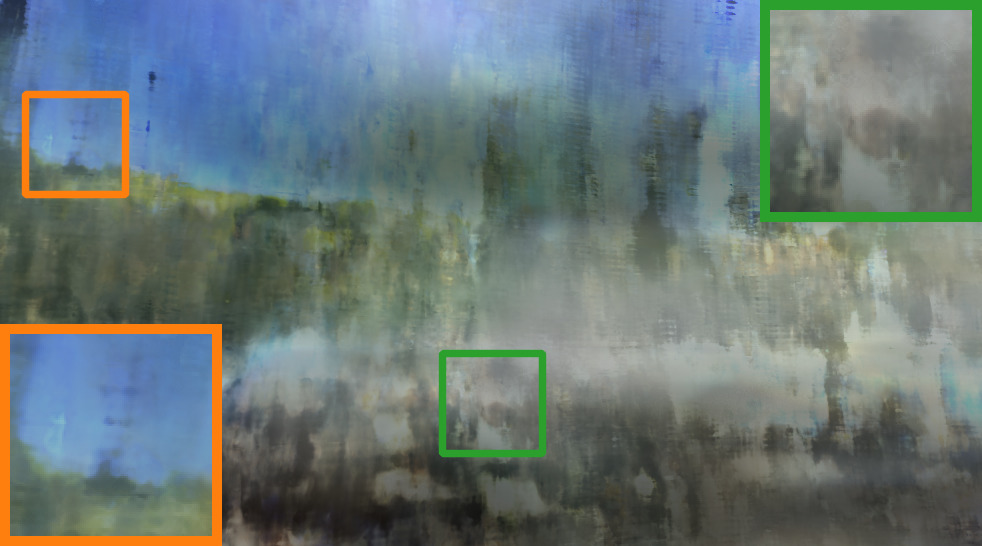} } & \raisebox{-0.5\height}{ \includegraphics[width=0.3167\linewidth]{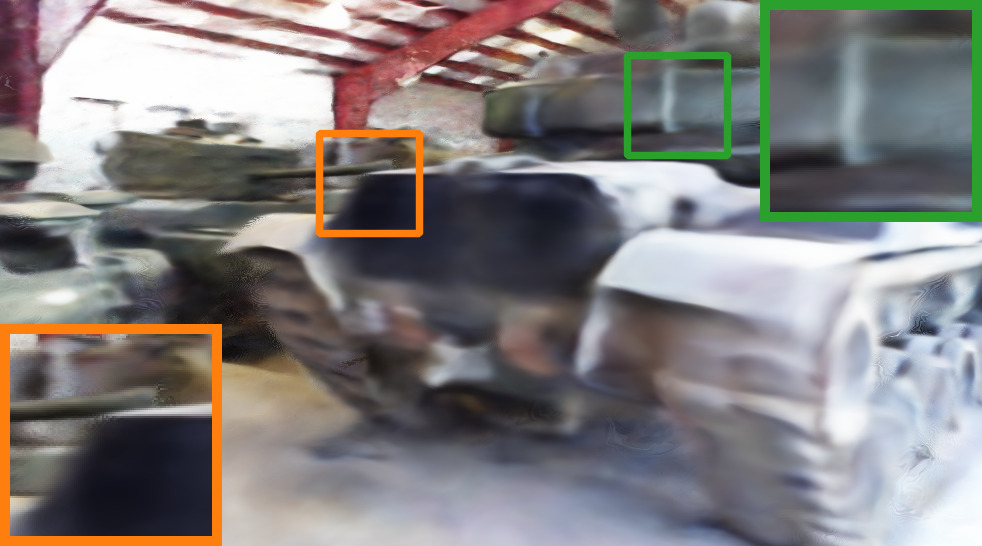} }\vspace{0.25em}\\
\rotatebox[origin=c]{90}{NPBG \cite{Aliev2020Neural}} & \raisebox{-0.5\height}{ \includegraphics[width=0.3167\linewidth]{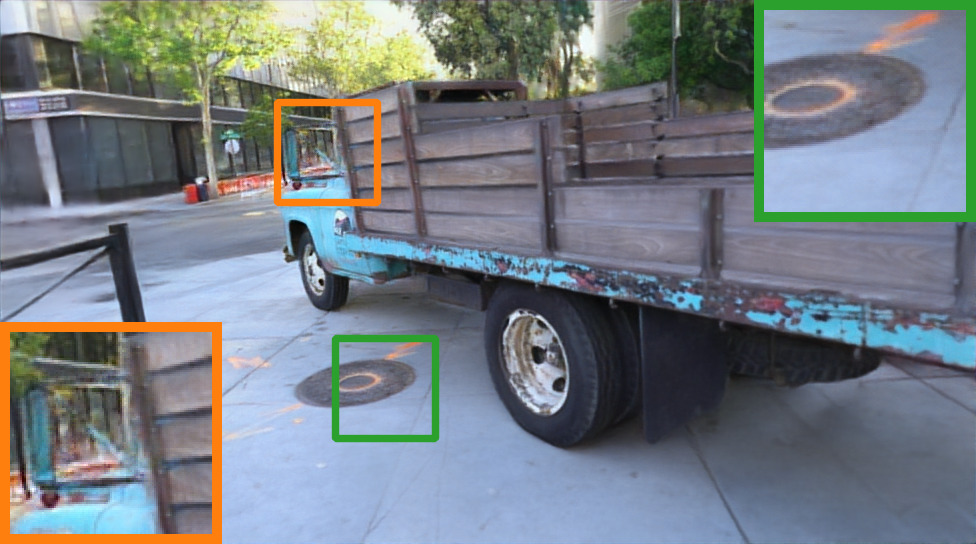} } & \raisebox{-0.5\height}{ \includegraphics[width=0.3167\linewidth]{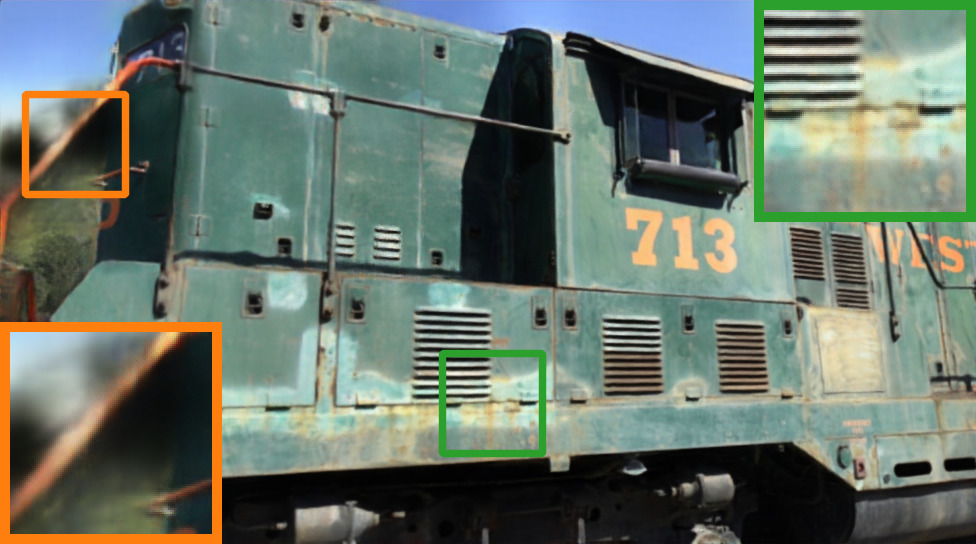} } & \raisebox{-0.5\height}{ \includegraphics[width=0.3167\linewidth]{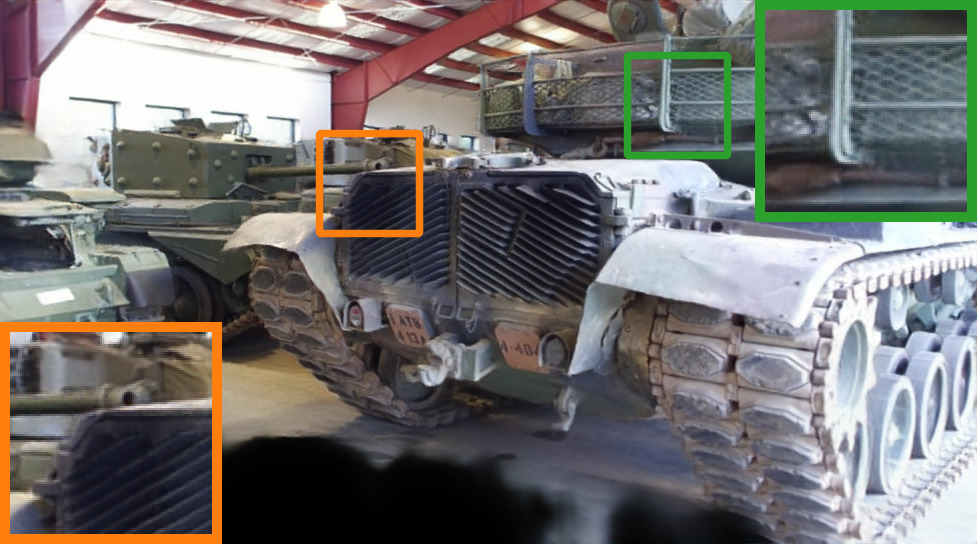} }\vspace{0.25em}\\
\rotatebox[origin=c]{90}{Ours} & \raisebox{-0.5\height}{ \includegraphics[width=0.3167\linewidth]{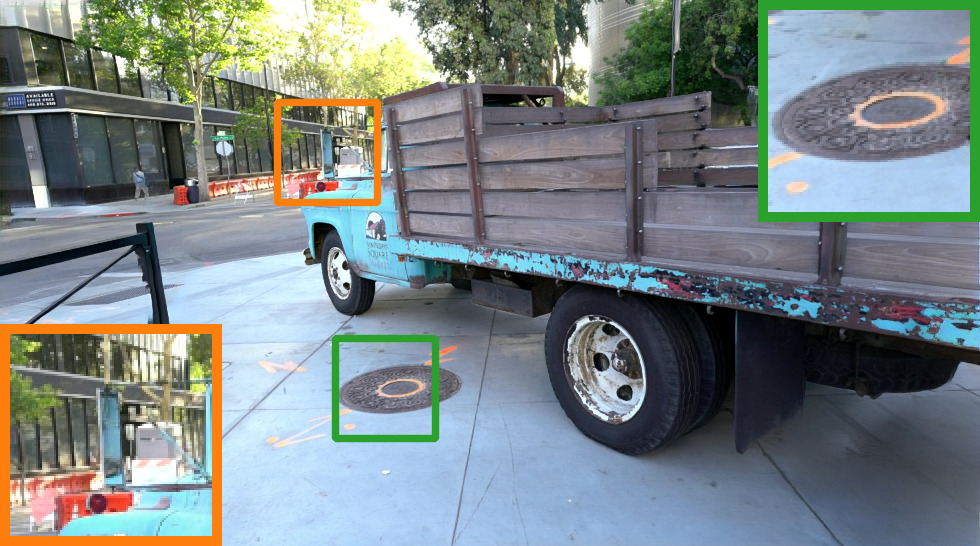} } & \raisebox{-0.5\height}{ \includegraphics[width=0.3167\linewidth]{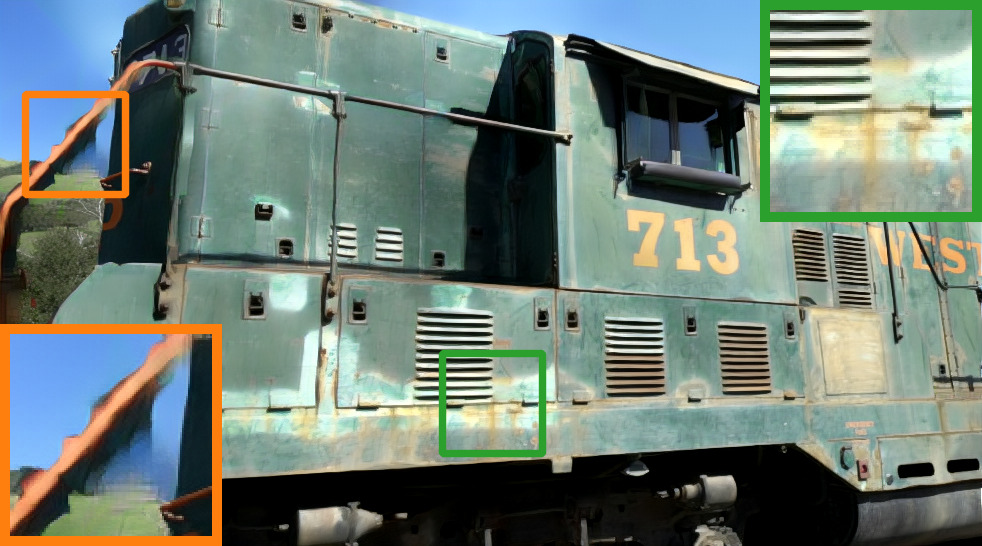} } & \raisebox{-0.5\height}{ \includegraphics[width=0.3167\linewidth]{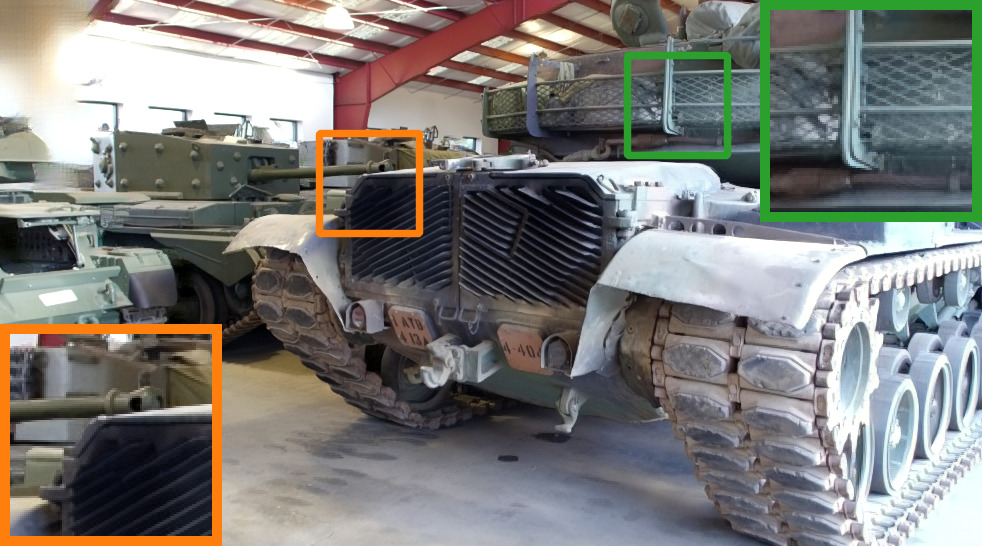} }\vspace{0.25em}\\
 & Truck & Train & M60\\
\end{tabular}
\endgroup

    \captionof{figure}{
        Qualitative results on Tanks and Temples. (Whole sequences withheld.)
    }
    \vspace{-4mm}
    \label{fig:tat_subseq}
\end{figure}

\myparagraph{Tanks and Temples.}
In this evaluation, we compare our approach to state-of-the-art methods on novel view sequences extracted from Tanks and Temples~\cite{Knapitsch2017Tanks}.
As we want to evaluate novel view synthesis from unstructured source images, we manually select a subset of camera poses from the test scenes as targets that we want to reconstruct.
These target images are taken out of the dataset and only serve as ground truth for evaluation of our synthesized results.
The scenes we use for evaluation have never been seen during the training of our method.
An example of the setup for the \emph{Truck} scene is depicted in Figure~\ref{fig:teaser}.

We compare our method to two recent state-of-the-art and two concurrent methods.
Extreme View Synthesis (\emph{EVS})~\cite{Choi2019Extreme} mainly focuses on extreme stereo baseline magnification and utilizes the multi-view stereo network MVSNet~\cite{Yao2018Mvsnet}.
Specifically, it warps the 3D feature volumes of the source images into the target view and fuses them.
We utilize the code provided by the authors and as no training code is available, we also apply the pretrained model that is provided.
Local Light Field Fusion (\emph{LLFF})~\cite{Mildenhall2019Local} is based on the multi-plane image idea and assumes that the poses of the source images lie on a plane.
For this method as well, we use the code provided by the authors and the provided pretrained model weights as no training code is available.

We also compare to Neural Radiance Fields (\emph{NeRF}), which is concurrent work that is published alongside ours~\cite{Mildenhall2020Nerf}.
Finally, we benchmark Neural Point-Based Graphics (\emph{NPBG}), which is likewise a concurrent publication~\cite{Aliev2020Neural}.
We train the descriptors per 3D point and fine-tune the provided rendering network per scene, utilizing the available code.
Note that \emph{NeRF} and \emph{NPBG} have to be trained on the test scenes, whereas our approach does not need any adaptation or fine-tuning on new scenes.

Quantitative results are summarized in Table~\ref{tab:tat_seq} and qualitative results are shown in Figure~\ref{fig:tat_subseq} and the supplementary video.
\emph{LLFF} clearly fails in this challenging unstructured setting: The assumptions of \emph{LLFF} are not met, which leads to strong ghosting artifacts.
\emph{EVS} works better, but often fails on fine details and sometimes misses whole parts of the image, although we used the very same set of source images as input as we selected for our method.
The latter artifacts are the main reason for the low quantitative performance of \emph{EVS}.
\emph{NeRF} struggles on the Tanks and Temples scenes.
The results are either blurry or fail completely, for example on the \emph{Train} scene.
\emph{NPBG} produces good images that are competitive with ours.
Our method produces sharp details and is superior to all other methods is terms of the LPIPS metric.

 \begin{figure}[t]
     \centering
     \begingroup
\setlength{\tabcolsep}{-0.1em}
\renewcommand{\arraystretch}{1}
\begin{tabular}{c c c c}
\rotatebox[origin=c]{90}{GT} & \raisebox{-0.5\height}{ \includegraphics[width=0.3167\linewidth]{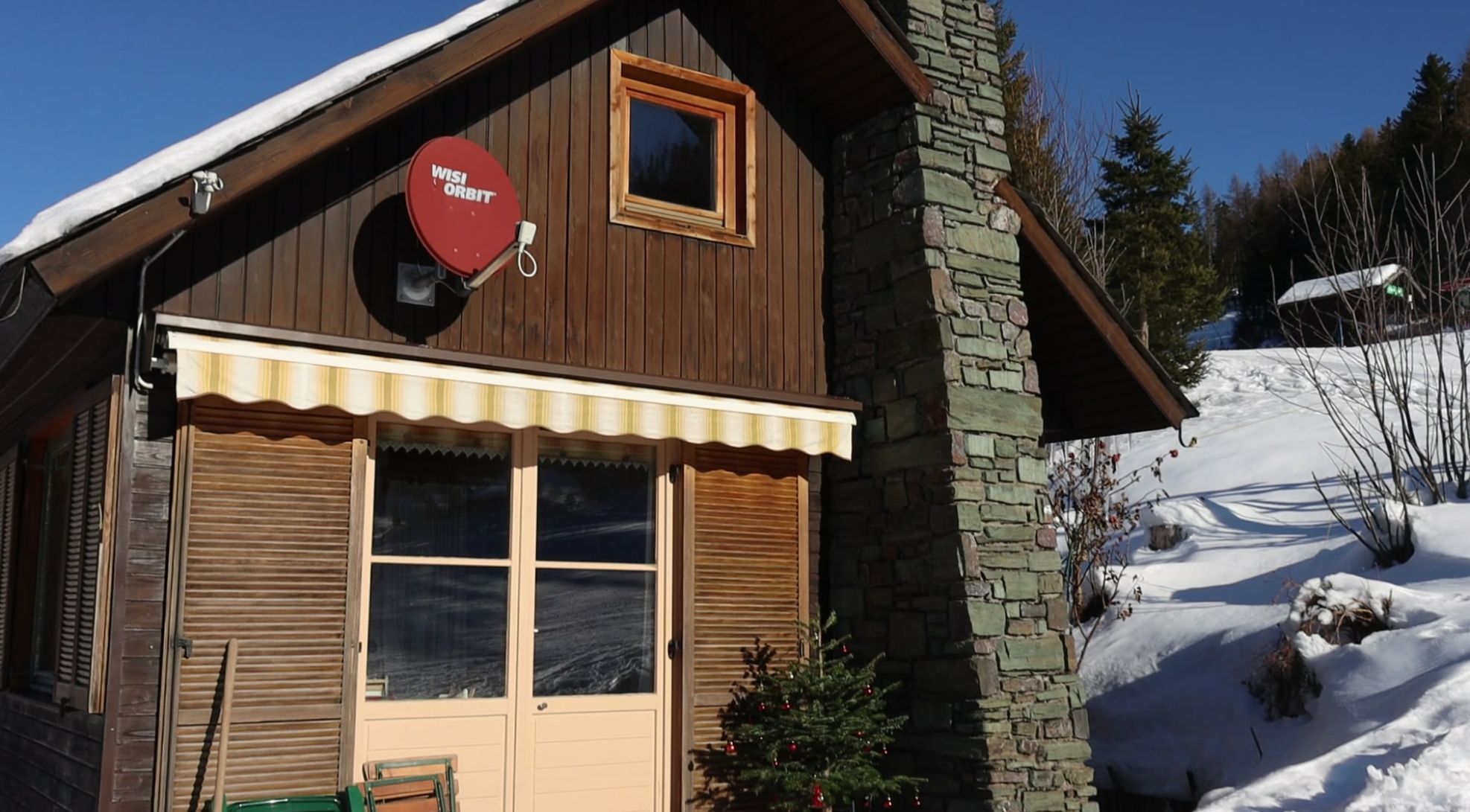} } & \raisebox{-0.5\height}{ \includegraphics[width=0.3167\linewidth]{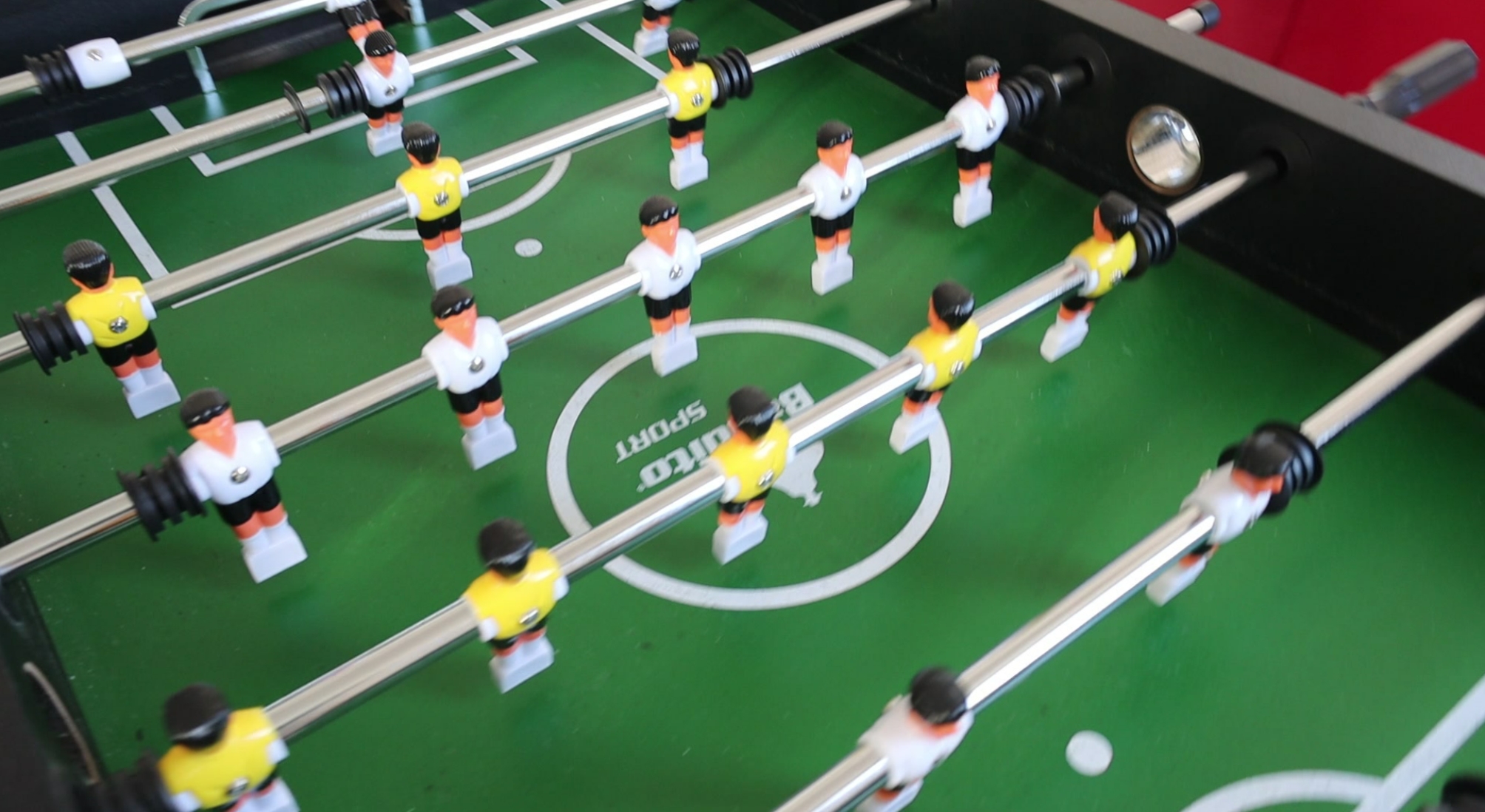} } & \raisebox{-0.5\height}{ \includegraphics[width=0.3167\linewidth]{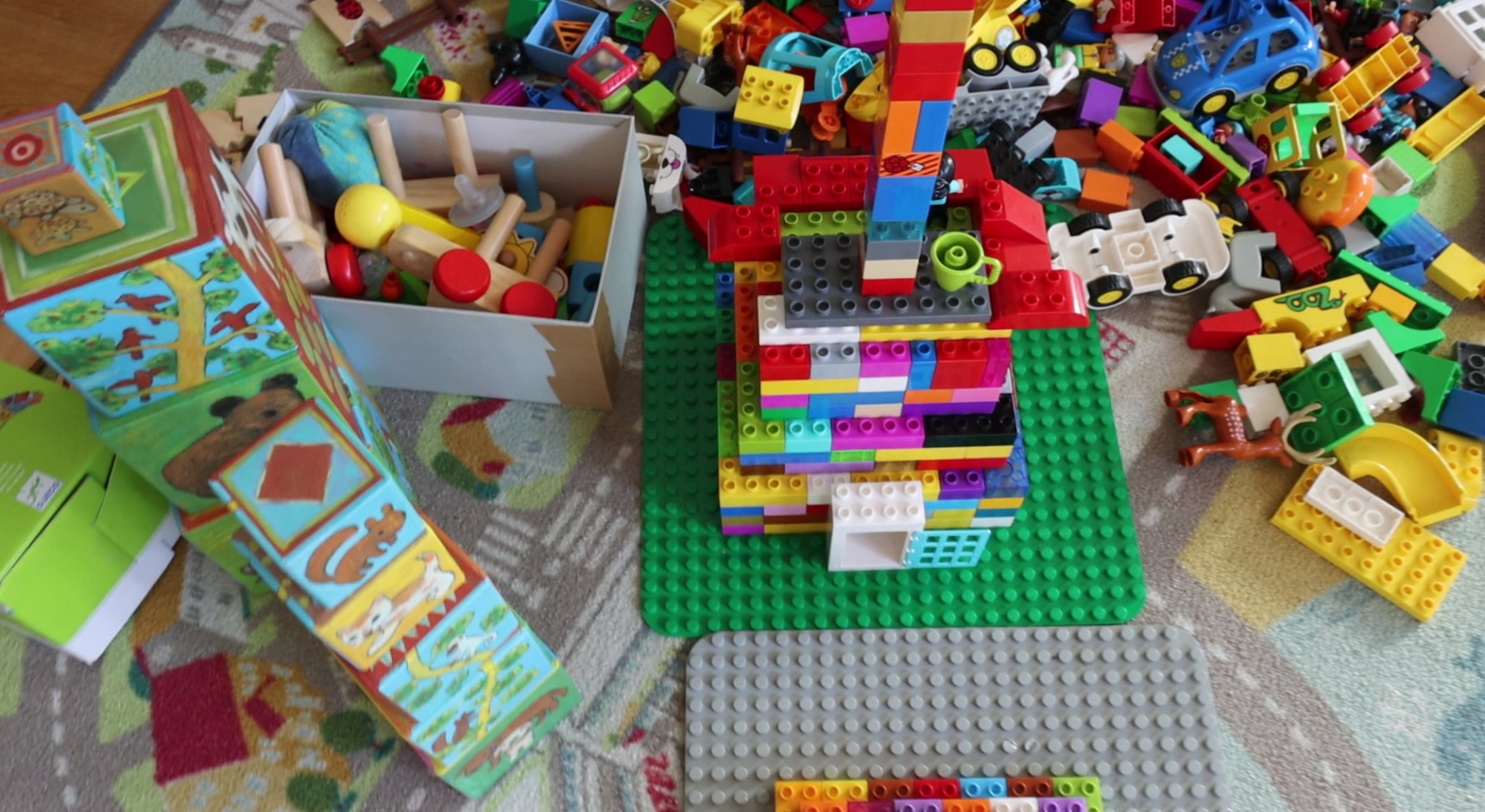} }\vspace{0.25em}\\
\rotatebox[origin=c]{90}{Ours} & \raisebox{-0.5\height}{ \includegraphics[width=0.3167\linewidth]{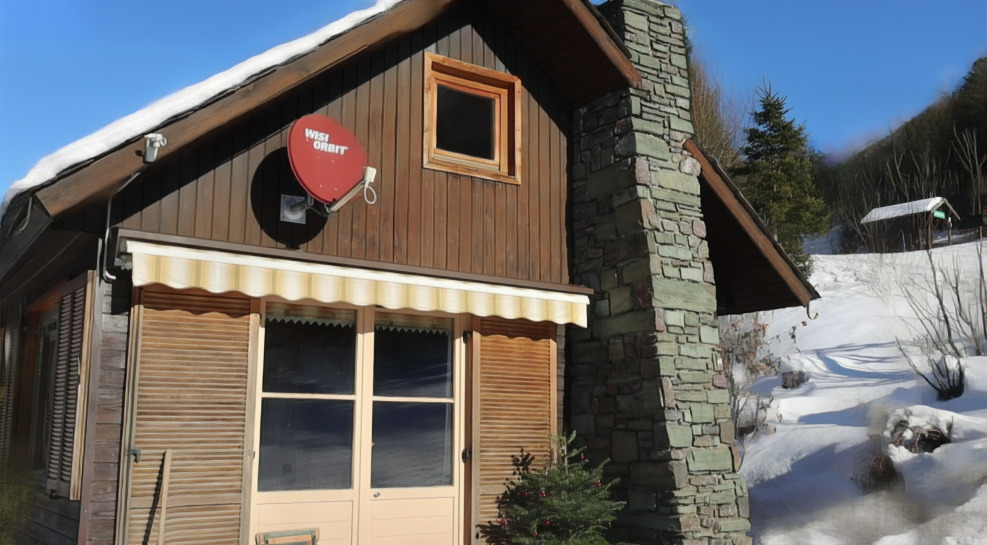} } & \raisebox{-0.5\height}{ \includegraphics[width=0.3167\linewidth]{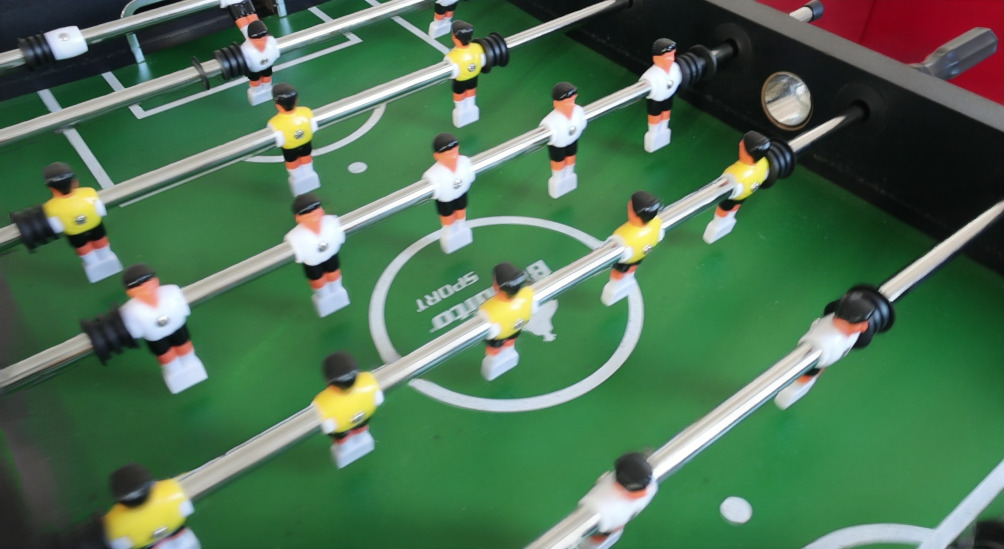} } & \raisebox{-0.5\height}{ \includegraphics[width=0.3167\linewidth]{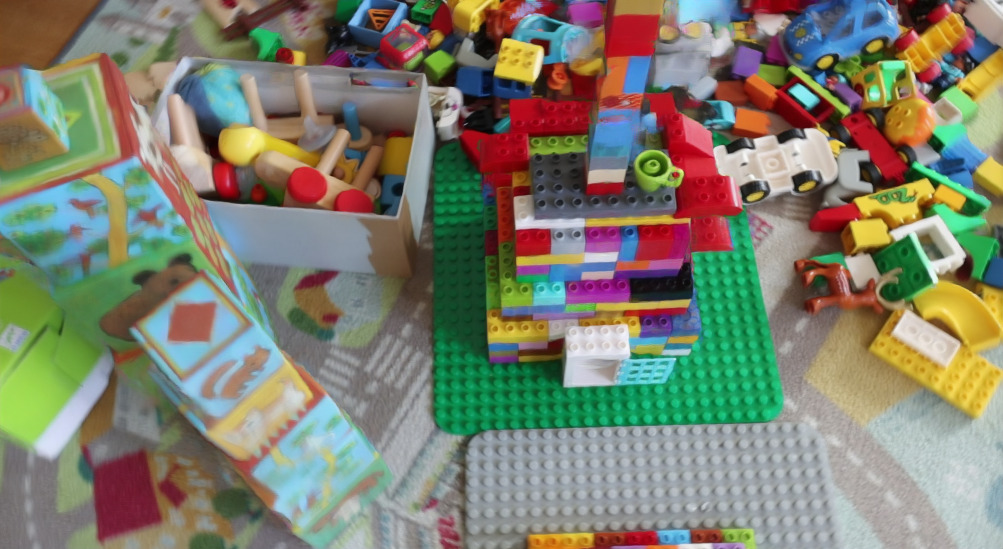} }\vspace{0.25em}\\
\end{tabular}
\endgroup

     \caption{
         Qualitative results on new recordings.
     }
     \label{fig:own}
     \vspace{-2em}
 \end{figure}

\myparagraph{New recordings.}
We also evaluate the presented method on new recordings that stress the unstructured setting.
We use a handheld camera in natural motion and record videos of different scenes to extract source images.
We then record each scene again to gather ground-truth data for new target views.
Results are shown in Figure~\ref{fig:own} and the supplementary video.

\begin{wrapfigure}[16]{r}{0.3\textwidth}
    \centering
    \vspace{-3mm}
    \includegraphics[width=0.99\linewidth]{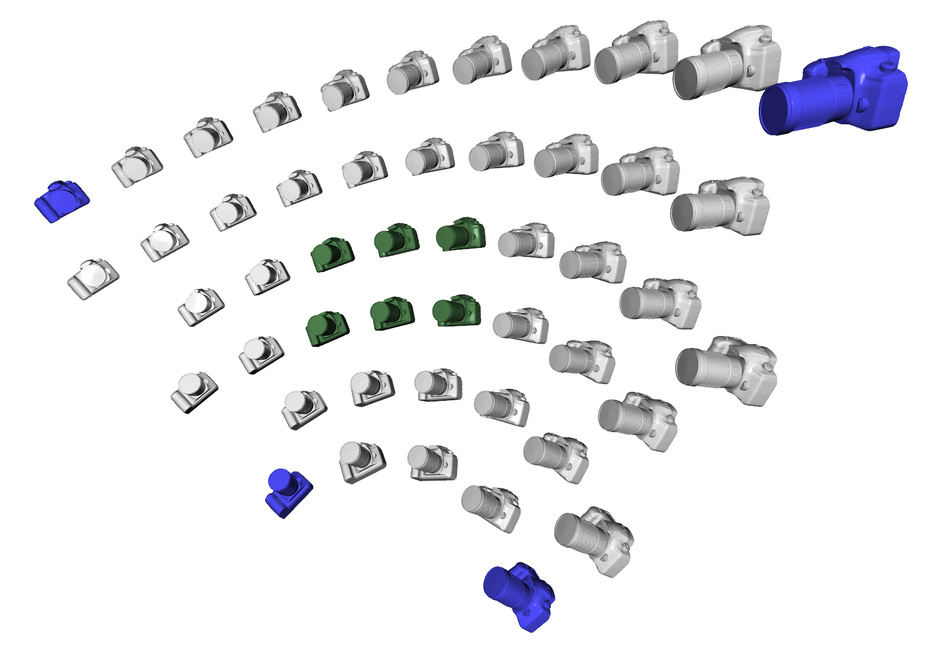}
    \caption{
        DTU evaluation setup.
        Gray cameras denote the source views.
        Green and blue cameras denote interpolation and extrapolation poses, respectively.
    }
    \label{fig:dtu_cams}
\end{wrapfigure}

\myparagraph{DTU.}
We now compare our method to state-of-the-art alternatives in a more constrained view interpolation and extrapolation setting.
For this we use the DTU dataset~\cite{Aanaes2016DTU}, which includes over $100$ tabletop scenes.
The image poses are identical for all scenes, as the camera has been mounted on a robotic arm and the views roughly cover an octant of a sphere.
Figure~\ref{fig:dtu_cams} visualizes the poses.

Of the $49$ camera poses we selected $10$ as targets for novel view synthesis and used the rest as source images.
We test the view extrapolation capabilities of all methods by having four target views on the corners of the camera grid. Interpolation is tested on $6$ center views.
We use the object masks for scenes $65$, $106$, and $118$, which are provided by \cite{Niemeyer2019Differentiable} for all source images.

\begin{table}[t]
    \centering
    \caption{
        Quantitative results on the DTU dataset.
        Numbers on the left are for view interpolation, numbers on the right are for extrapolation.
    }
    \resizebox{1\linewidth}{!}{
    \scriptsize
    \begin{tabular}{rrrrrrrrrr}
\toprule
\multicolumn{1}{c}{} & \multicolumn{3}{c}{Scan 65} & \multicolumn{3}{c}{Scan 106} & \multicolumn{3}{c}{Scan 118} \\
\cmidrule(lr){2-4}
\cmidrule(lr){5-7}
\cmidrule(lr){8-10}
 & \multicolumn{1}{c}{$\downarrow$LPIPS} & \multicolumn{1}{c}{$\uparrow$SSIM} & \multicolumn{1}{c}{$\uparrow$PSNR} & \multicolumn{1}{c}{$\downarrow$LPIPS} & \multicolumn{1}{c}{$\uparrow$SSIM} & \multicolumn{1}{c}{$\uparrow$PSNR} & \multicolumn{1}{c}{$\downarrow$LPIPS} & \multicolumn{1}{c}{$\uparrow$SSIM} & \multicolumn{1}{c}{$\uparrow$PSNR} \\
\midrule
\multicolumn{1}{l}{EVS \cite{Choi2019Extreme}} & 0.61/0.53 & 0.938/0.917 & 23.07/21.23 & 0.75/0.53 & 0.903/0.880 & 19.95/18.62 & 0.47/0.42 & 0.931/0.911 & 23.00/20.47 \\
\multicolumn{1}{l}{LLFF \cite{Mildenhall2019Local}} & 0.51/0.44 & 0.939/0.926 & 22.44/22.04 & 0.61/0.39 & 0.907/0.893 & 24.08/24.61 & 0.47/0.30 & 0.932/0.929 & 28.95/27.40 \\
\multicolumn{1}{l}{NeRF \cite{Mildenhall2020Nerf}} & {\bf 0.17}/0.32 & {\bf 0.987}/{\bf 0.963} & {\bf 34.41}/{\bf 27.81} & 0.36/0.40 & {\bf 0.973}/0.931 & {\bf 34.52}/24.36 & 0.24/0.27 & {\bf 0.985}/{\bf 0.952} & {\bf 37.16}/{\bf 28.39} \\
\multicolumn{1}{l}{NPBG \cite{Aliev2020Neural}} & 0.82/0.96 & 0.896/0.839 & 17.77/15.59 & 0.94/0.53 & 0.856/0.879 & 20.70/22.54 & 0.74/0.41 & 0.876/0.905 & 24.10/24.97 \\
\multicolumn{1}{l}{Our} & 0.25/{\bf 0.30} & 0.972/0.950 & 26.96/24.08 & {\bf 0.25}/{\bf 0.26} & 0.963/{\bf 0.938} & 27.24/{\bf 24.63} & {\bf 0.16}/{\bf 0.20} & 0.975/0.951 & 29.21/25.75 \\
\bottomrule
\end{tabular}
    }
    \vspace{-2em}
    \label{tab:dtu}
\end{table}

\begin{figure}[t!]
    \centering
    \resizebox{1\linewidth}{!}{
    \begingroup
\setlength{\tabcolsep}{0.1em}
\renewcommand{\arraystretch}{1}
\begin{tabular}{c c c c c c}
\rotatebox[origin=c]{90}{GT} & \raisebox{-0.5\height}{ \includegraphics[width=0.3000\linewidth]{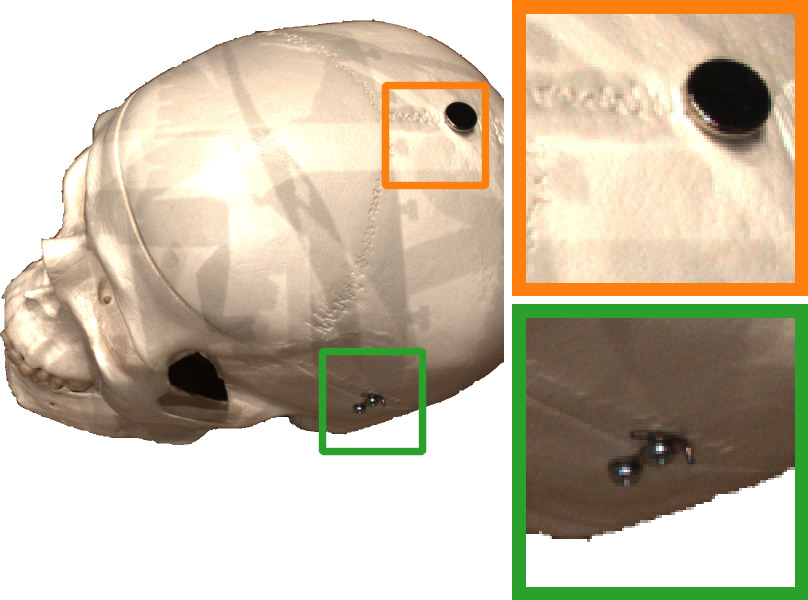} } & \raisebox{-0.5\height}{ \includegraphics[width=0.3000\linewidth]{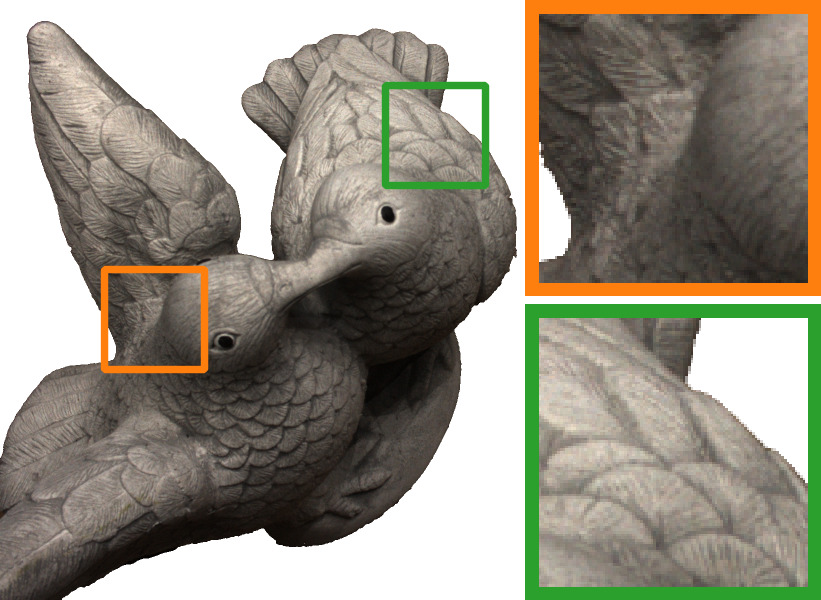} } & \raisebox{-0.5\height}{ \includegraphics[width=0.3000\linewidth]{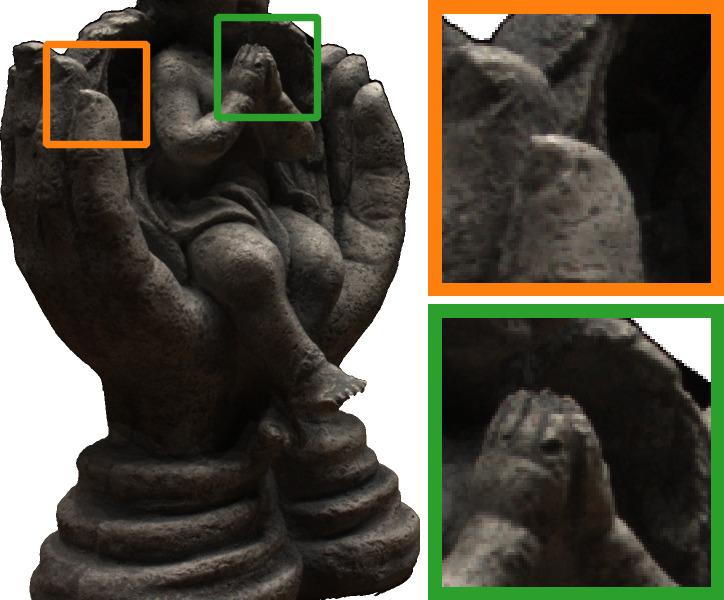} }\vspace{0.25em}\\
\rotatebox[origin=c]{90}{EVS \cite{Choi2019Extreme}} & \raisebox{-0.5\height}{ \includegraphics[width=0.3000\linewidth]{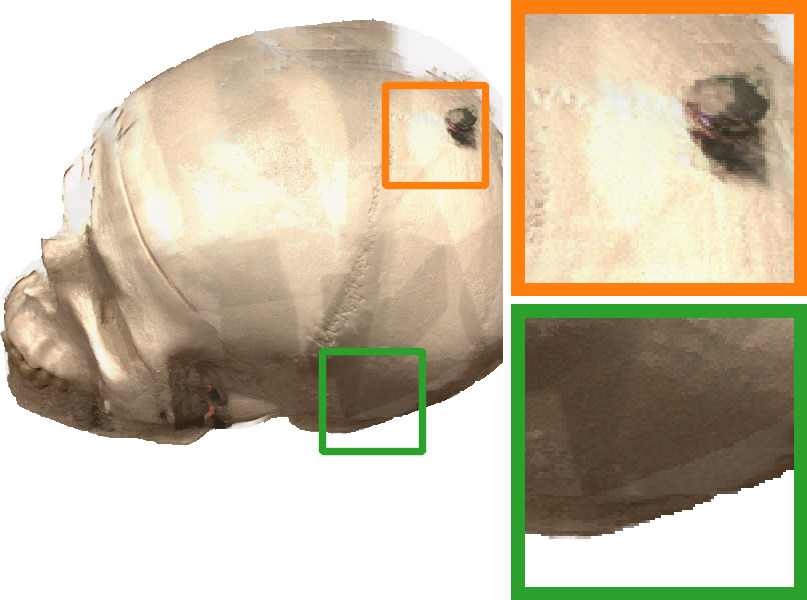} } & \raisebox{-0.5\height}{ \includegraphics[width=0.3000\linewidth]{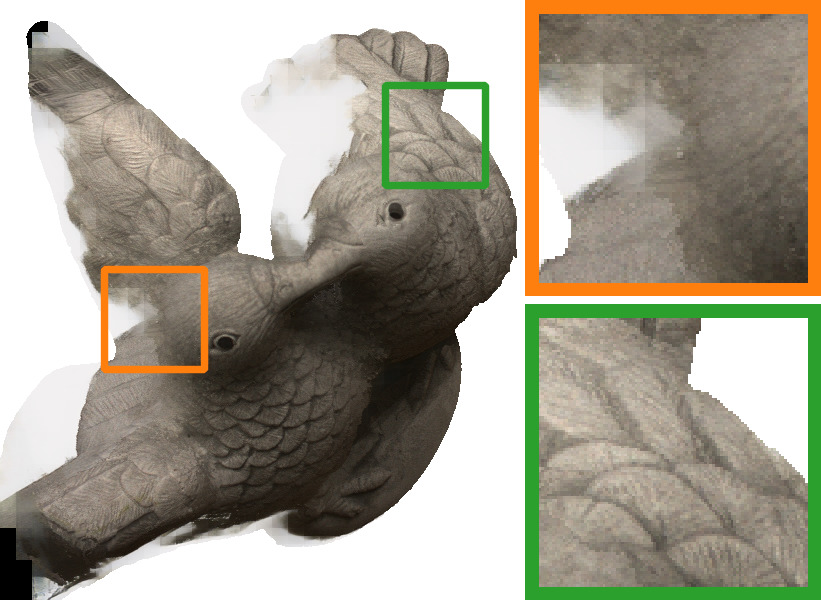} } & \raisebox{-0.5\height}{ \includegraphics[width=0.3000\linewidth]{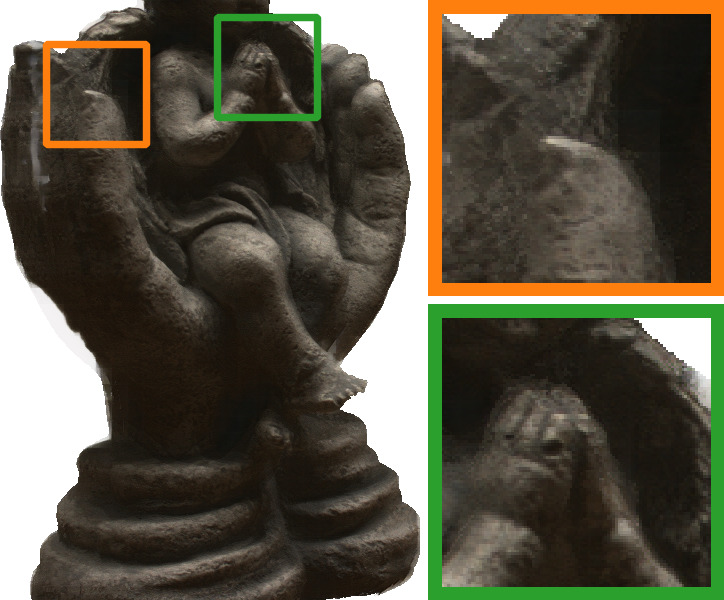} }\vspace{0.25em}\\
\rotatebox[origin=c]{90}{LLFF \cite{Mildenhall2019Local}} & \raisebox{-0.5\height}{ \includegraphics[width=0.3000\linewidth]{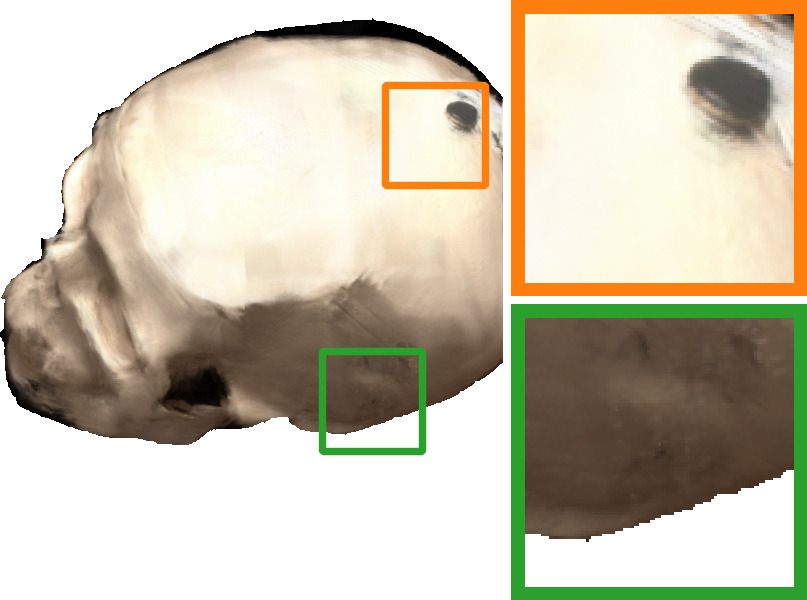} } & \raisebox{-0.5\height}{ \includegraphics[width=0.3000\linewidth]{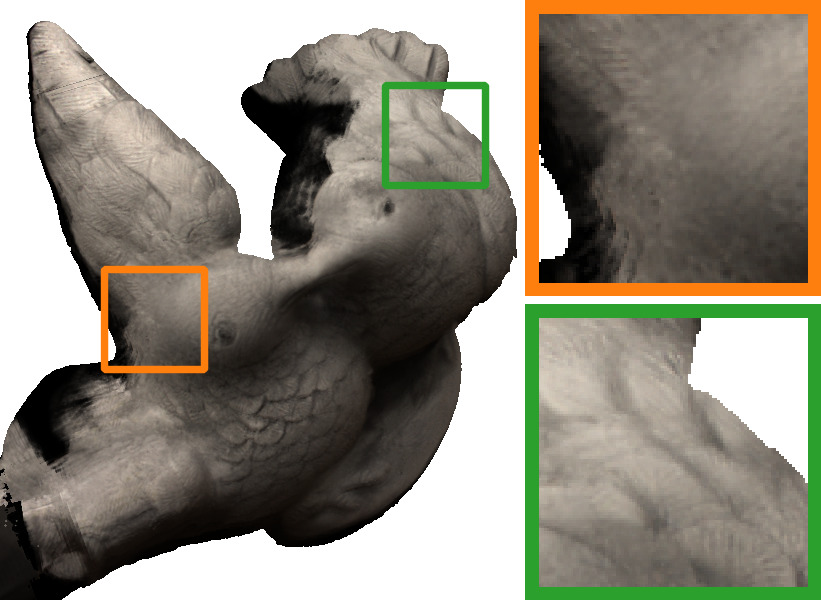} } & \raisebox{-0.5\height}{ \includegraphics[width=0.3000\linewidth]{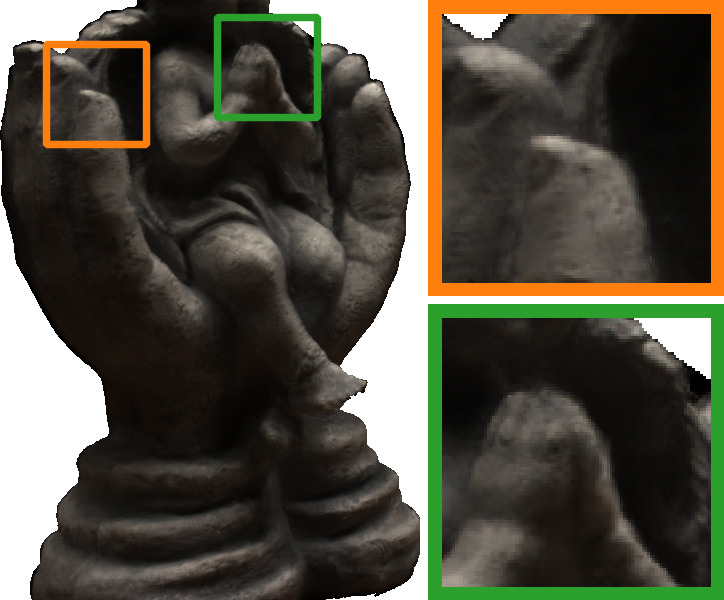} }\vspace{0.25em}\\
\rotatebox[origin=c]{90}{NeRF \cite{Mildenhall2020Nerf}} & \raisebox{-0.5\height}{ \includegraphics[width=0.3000\linewidth]{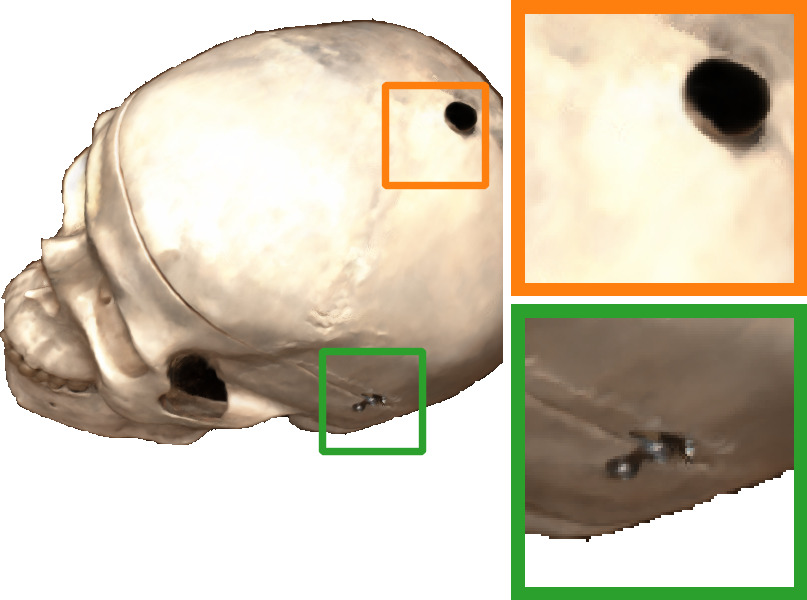} } & \raisebox{-0.5\height}{ \includegraphics[width=0.3000\linewidth]{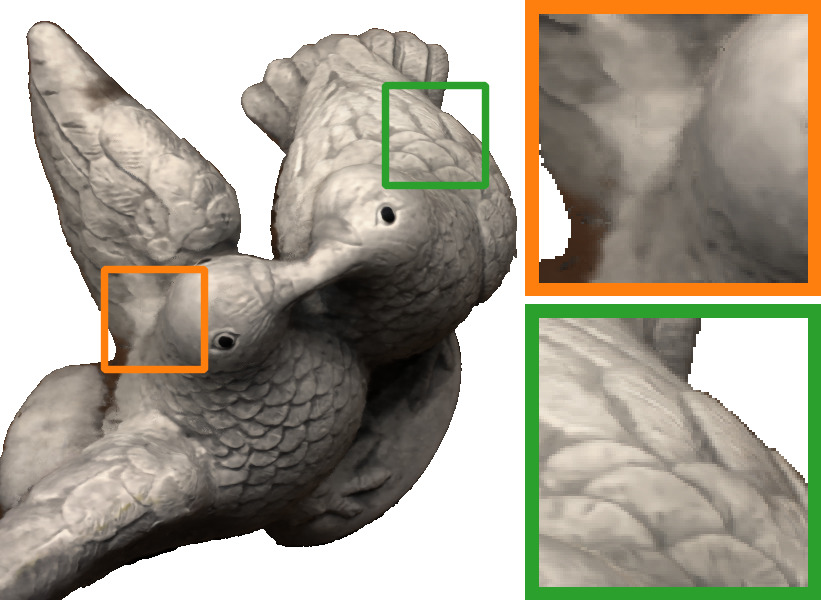} } & \raisebox{-0.5\height}{ \includegraphics[width=0.3000\linewidth]{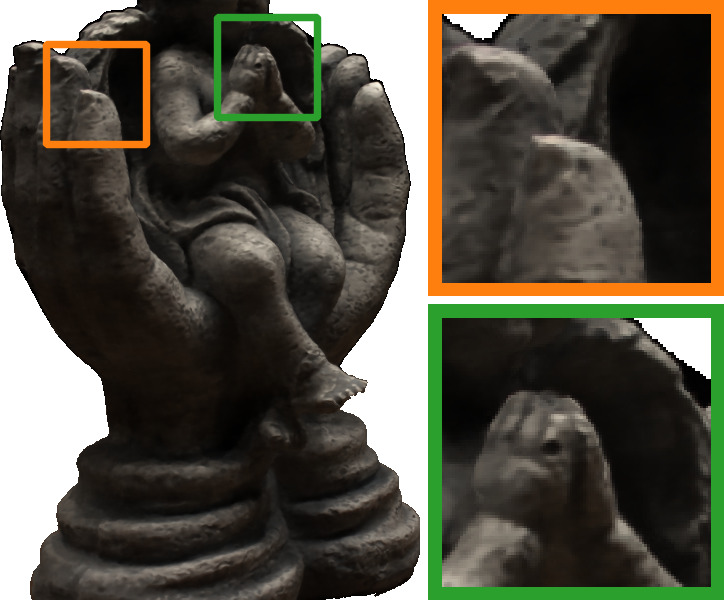} }\vspace{0.25em}\\
\rotatebox[origin=c]{90}{NPBG \cite{Aliev2020Neural}} & \raisebox{-0.5\height}{ \includegraphics[width=0.3000\linewidth]{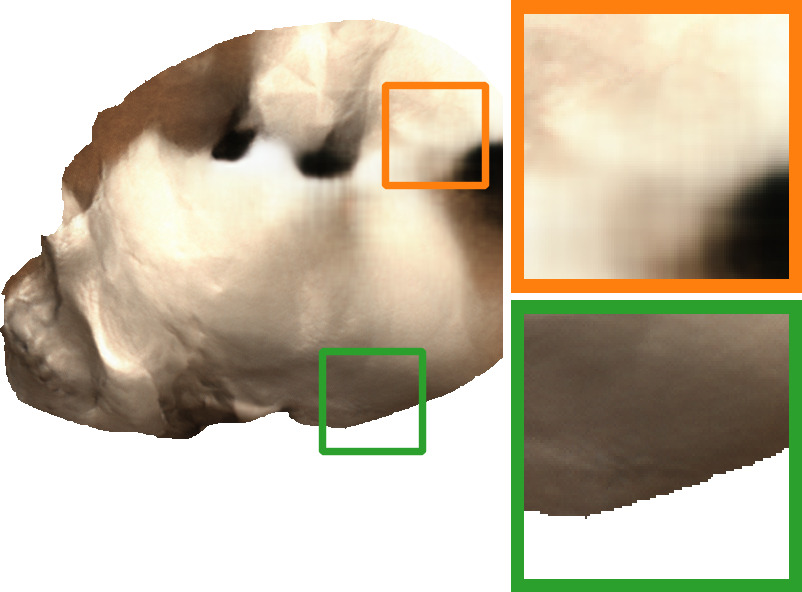} } & \raisebox{-0.5\height}{ \includegraphics[width=0.3000\linewidth]{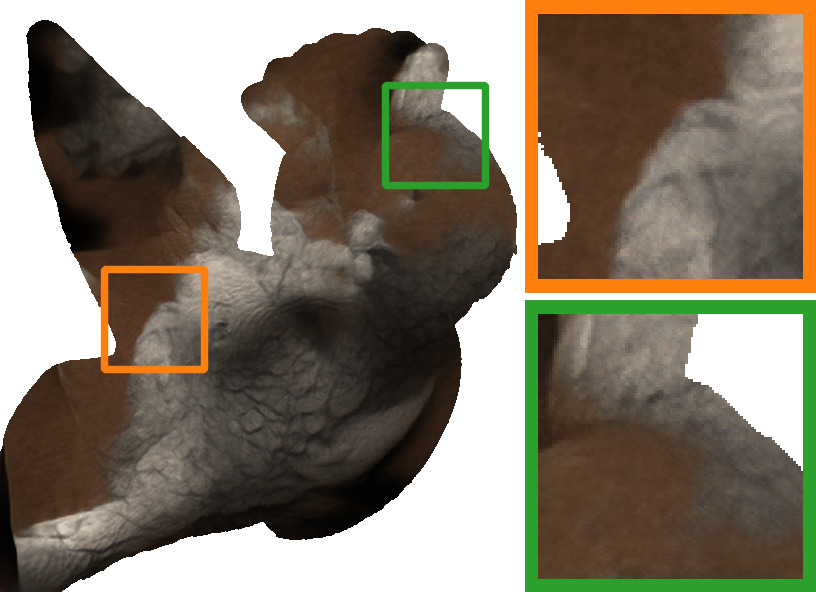} } & \raisebox{-0.5\height}{ \includegraphics[width=0.3000\linewidth]{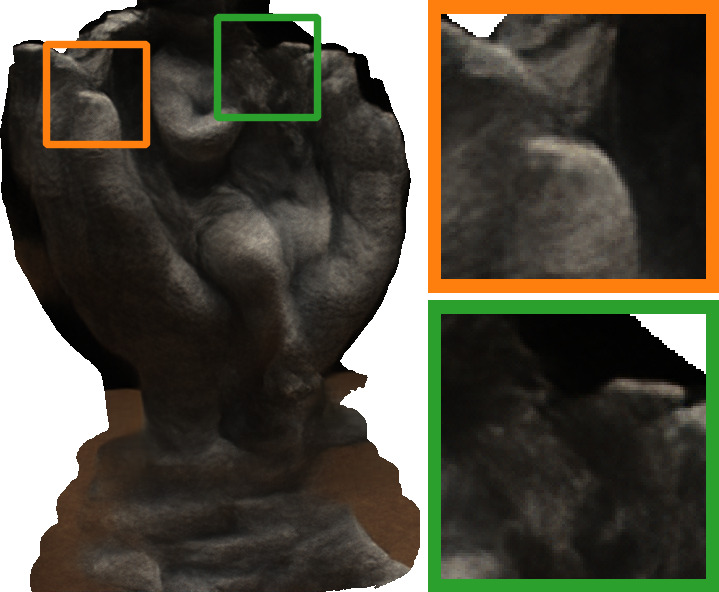} }\vspace{0.25em}\\
\rotatebox[origin=c]{90}{Ours} & \raisebox{-0.5\height}{ \includegraphics[width=0.3000\linewidth]{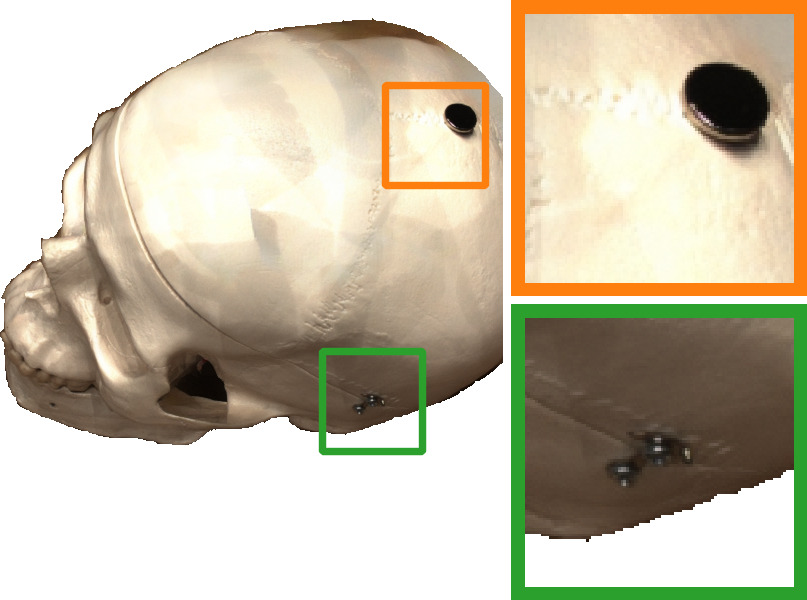} } & \raisebox{-0.5\height}{ \includegraphics[width=0.3000\linewidth]{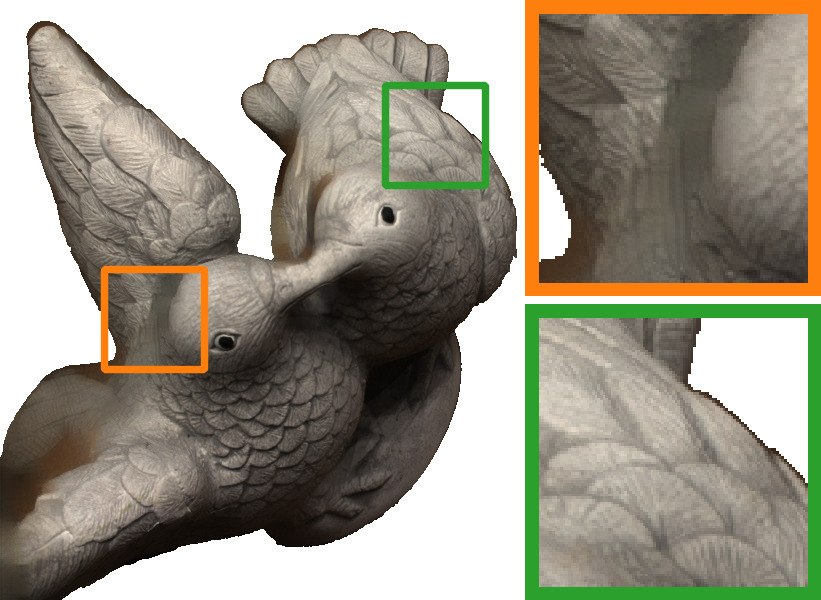} } & \raisebox{-0.5\height}{ \includegraphics[width=0.3000\linewidth]{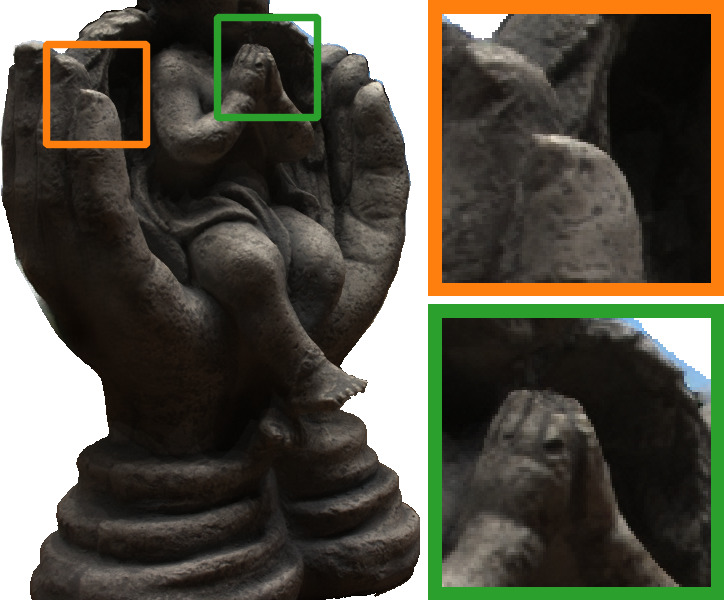} }\vspace{0.25em}\\
 & Scan 65 & Scan 106 & Scan 118\\
\end{tabular}
\endgroup

    }
    \caption{
        View extrapolation results on the DTU dataset.
    }
    \label{fig:dtu}
\end{figure}

We summarize the quantitative results in Table~\ref{tab:dtu}.
Qualitative results are shown in Figure~\ref{fig:dtu} and the supplementary video.
We notice blending artifacts in the images produced by \emph{EVS}~\cite{Choi2019Extreme}, which are reflected in the lower quantitative performance.
On the other hand, the results of \emph{LLFF}~\cite{Mildenhall2019Local} and especially of \emph{NeRF}~\cite{Mildenhall2020Nerf} are clearly better compared to their performance on Tanks and Temples.
This comes as no surprise because the DTU setup is much closer to the basic assumptions of these methods: the scene is a clearly bounded object and the camera poses are densely and regularly sampled by the source views.
\emph{NPBG}~\cite{Aliev2020Neural} struggles in this setting and often intermixes background and foreground.
In contrast, our method yields sharp results, including in the extrapolation setting, and performs reasonable inpainting when geometry is missing.
Note that the illumination varies with the viewing direction in DTU scenes.
While we are still able to synthesize realistic results, low-level metrics such as PSNR are not very reliable.

\myparagraph{Limitations.}
While our method is a clear step forward compared to prior work, it has limitations.
The first limitation is apparent when we examine videos rendered from a sequence of new views.
We only synthesize images frame-by-frame and do not enforce any temporal consistency. Thus synthesized videos exhibit temporal instability.
The second limitation stems from the use of proxy 3D geometry.
If the 3D model used for mapping misses large parts of the scene or has gross outliers, our pipeline will produce visible artifacts.
The flip side is that our approach can directly benefit from future improvements in SfM and MVS pipelines~\cite{Knapitsch2017Tanks}.

\section{Conclusion}
We presented a method for novel view synthesis in the challenging setting of unstructured input images acquired by natural motion through the scene.
After preprocessing the input using standard SfM and MVS to get camera parameters and 3D proxy geometry, we showed that a recurrent mapping and blending architecture can produce sharp images for new views of the scene that depart significantly from the input.
The recurrent architecture enables using an arbitrary number of source images per target view, which mostly eliminates the need for hand-crafted heuristics for source image selection and demonstrably helps in the unstructured setting.
In future work, we plan to improve temporal consistency.
We also expect that the results of our method will continue to improve as new techniques for SfM, MVS, and surface reconstruction are introduced.

\myparagraph{Acknowledgements.} We thank Kai Zhang for the evaluation of NeRF.

\clearpage
\bibliographystyle{splncs04}
\bibliography{egbib_longstr,egbib}
\end{document}